\documentclass[lettersize,journal]{IEEEtran}
\usepackage{amsmath,amsfonts}
\usepackage[noend]{algorithmic}
\usepackage{algorithm}
\usepackage{booktabs}
\usepackage{tabularx}
\usepackage{float}
\usepackage{hyperref}
\usepackage{array}
\usepackage[caption=false,font=normalsize,labelfont=sf,textfont=sf]{subfig}
\usepackage{textcomp}
\usepackage{stfloats}
\usepackage{url}
\makeatletter
\g@addto@macro{\UrlBreaks}{\do\/\do\-\do\a\do\b\do\c\do\d\do\e\do\f\do\g\do\h\do\i\do\j\do\k\do\l\do\m\do\n\do\o\do\p\do\q\do\r\do\s\do\t\do\u\do\v\do\w\do\x\do\y\do\z\do\A\do\B\do\C\do\D\do\E\do\F\do\G\do\H\do\I\do\J\do\K\do\L\do\M\do\N\do\O\do\P\do\Q\do\R\do\S\do\T\do\U\do\V\do\W\do\X\do\Y\do\Z\do\0\do\1\do\2\do\3\do\4\do\5\do\6\do\7\do\8\do\9}
\makeatother
\usepackage{verbatim}
\usepackage{graphicx}
\usepackage{cite}
\usepackage[justification=centering]{caption}
\begin{document}

\title{SCOPE-FL: A Strategy-proof Chain-based Optimal pareto efficient Federated Learning System}

\author{Seyed~Salar~Ghazi,~\IEEEmembership{Student~Member,~IEEE,}
        Kaiwen~Zhang,~\IEEEmembership{Member,~IEEE,}
        Mehdi~Feizi,
        and~Hans-Arno~Jacobsen,~\IEEEmembership{Fellow,~IEEE}
        \thanks{Seyed Salar Ghazi and Kaiwen Zhang are with the Department of Software and IT Engineering, École de Technologie Supérieure (ÉTS), Montréal, QC H3C 1K3, Canada (e-mail: seyed-salar.ghazi.1@ens.etsmtl.ca; kaiwen.zhang@etsmtl.ca).}
\thanks{Mehdi Feizi is with the Department of Economics, Faculty of Economics and Administrative Sciences, Ferdowsi University of Mashhad, Mashhad, Iran (e-mail: feizi@um.ac.ir).}
\thanks{Hans-Arno Jacobsen is with the Department of Electrical and Computer Engineering, University of Toronto, Toronto, ON M5S 3G4, Canada (e-mail: jacobsen@eecg.toronto.edu).}}

% \IEEEpubid{0000--0000/00\$00.00~\copyright~2021 IEEE}
% Remember, if you use this you must call \IEEEpubidadjcol in the second
% column for its text to clear the IEEEpubid mark.

\maketitle

\begin{abstract}
Hierarchical Federated Learning (HFL) enables
scalable collaborative model training across distributed devices while preserving data privacy. However, existing HFL client selection mechanisms suffer from a fundamental strategic inefficiency. By prioritizing stability over Pareto efficiency (PE), they produce suboptimal resource allocations, and without strategy proofness (SP), participants are incentivized to misrepresent their true preferences, both failures degrading system overall welfare in the Pareto sense in practice. To address it, we propose SCOPE-FL (Strategy-proof Chain-based Optimal pareto efficient Federated Learning), a synchronous HFL framework that formulates client selection as a two-sided school choice problem solved through the Top Trading Cycle (TTC) algorithm that simultaneously guarantees PE and SP. For reward distribution, SCOPE-FL employs a scalable Shapley value approximation based on One-Round Reconstruction (OR), ensuring compensation proportional to each client's contribution. The entire mechanism executes via blockchain smart contracts, providing the tamper-proof environment required for the SP guarantees to hold in practice. A comprehensive evaluation on MNIST, Fashion-MNIST, and CIFAR-10 demonstrates that SCOPE-FL outperforms state-of-the-art approaches, including DA, IAS, and other methods across model accuracy, convergence rate, and reward efficiency, while achieving communication latency comparable to DA and blockchain overhead significantly lower than DA at scale. 
\end{abstract}


\section{Introduction} \label{intro}
\IEEEPARstart{C}{ollaborative} machine learning across distributed devices raises fundamental challenges in communication efficiency, data privacy, and strategic inefficiency in client selection, where participants misrepresent their true preferences and existing mechanisms fail to ensure Pareto efficiency over reported preferences, failing to optimize client's welfare. Federated Learning (FL) addresses the first two concerns by enabling machine learning models to be trained collaboratively across many devices without requiring raw data to be shared. In each training round, selected clients receive the current global model, train it locally on their own data, and return model updates to a central server, which aggregates them into an improved global model. This iterative process continues until convergence. \cite{ayeelyan2024federated,jung2025federated}.

While single-server FL has proven effective for moderate-scale deployments, it introduces critical bottlenecks as the number of participating devices grows. A single central server represents a point of failure, a communication bottleneck, and a centralized target for security attacks. Hierarchical Federated Learning (HFL) addresses these structural limitations by introducing a layered architecture in which edge servers aggregate local updates from nearby clients before forwarding them to a principal server for global aggregation \cite{zuo2024spyker,wehbi2022towards}. This hierarchical design significantly reduces communication overhead, improves scalability, and enhances fault tolerance, making HFL particularly well-suited for geo-distributed environments where clients are widely dispersed and network conditions are highly variable \cite{zuo2024spyker,hudson2025flight}.

Despite these structural advantages, existing HFL systems suffer from a fundamental limitation in their client selection mechanisms, called strategic inefficiency, that prevents them from realizing their full potential, and they fail to achieve PE. The dominant design philosophy prioritizes stability, ensuring no client-server pair mutually prefers to deviate from their current assignment at the direct expense of Pareto efficiency (PE) \cite{wehbi2022towards, yellampalli2024client}. A stable matching is not generally pareto efficient, meaning the system accepts an allocation where aggregate utility across all participants could be improved without disadvantaging anyone. In welfare terms, stability-oriented mechanisms leave potential gains in model quality, resource utilization, and economic outcomes unrealized in every training round \cite{donahue2021optimality, bichler2017market, haeringer2018market}. For HFL systems operating across large client populations and multiple servers, this systematic welfare sacrifice compounds across rounds, cumulatively degrading both learning performance and participant satisfaction.

Pursuing PE alone, however, is insufficient to guarantee system welfare in practice. Without strategy proofness (SP), self-interested participants have incentives to misrepresent their true preferences (based on various parameters, including reward, latency, and capacity) and capabilities to obtain more favorable assignments. A client may misreport its computational costs to appear more attractive to high-reward servers. When participants misreport, the preference profile fed into the matching mechanism is corrupted, and the resulting assignment is pareto efficient only with respect to false inputs, which means it is welfare-destroying in reality and strategically inefficient. Realizing pareto efficient outcomes in practice therefore requires PE and SP not as independent goals but as inseparable properties. PE defines the welfare target, and SP is what ensures participants reveal the information needed to reach it \cite{bichler2017market, haeringer2018market}.

Existing approaches address pieces of this problem but none resolve it comprehensively. Stability-focused mechanisms such as Deferred Acceptance (DA) \cite{bichler2017market, wehbi2022towards, yellampalli2024client} guarantee stable matchings but, as noted above, sacrifice PE. pareto efficient mechanisms such as Immediate Acceptance with Skip (IAS) \cite{bichler2017market, haeringer2018market} achieve efficiency but lack SP entirely, leaving the system vulnerable to manipulation. Auction-based and contract-theoretic incentive mechanisms\cite{gupta2023federated, ali2023systematic} optimize server-side utility but do not provide SP guarantees. Reputation-based systems \cite{ali2023systematic} improve trust over time but cannot guarantee PE in any given round.

Beyond matching, evaluating the contributions of heterogeneous clients to the global model remains a persistent bottleneck because the methods, including Shapley value calculation, are combinatorially intractable at scale \cite{song2019profit, batool2023block}, while lightweight approximations either sacrifice accuracy in non-convex settings or neglect coalition effects among clients \cite{batool2023block}. Finally, even in seemingly decentralized HFL architectures, the matching process often remains centralized at a single principal server, creating single point of failure, manipulation, and bias \cite{wehbi2022towards, yellampalli2024client, wei2022participant}.

{\sloppy
To address these limitations, we propose SCOPE-FL (\textbf{S}trategy-proof \textbf{C}hain-based \textbf{O}ptimal \textbf{P}areto-\textbf{E}fficient \textbf{F}ederated \textbf{L}earning), a novel synchronous HFL framework that guarantees a pareto efficient allocation by formulating client selection as a two-sided school choice problem solved through the Top Trading Cycle (TTC) algorithm \cite{bichler2017market, haeringer2018market}. The school choice formulation naturally captures the bilateral nature of HFL client selection, where both clients and servers hold preferences over each other based on reward, latency, and contribution quality. Although HFL servers collaborate toward a single global model, the school choice analogy does not require them to be rivals. In the canonical formulation \cite{abdulkadirouglu2003school}, schools are not strategic agents but objects to be ”consumed” by students, holding only capacities and priorities, while students alone hold preferences and act strategically. Furthermore, each server rewards high-contribution clients; competing for desirable clients is equivalent to competing for those that most improve the shared model. 

This asymmetry does not mean that server interests are neglected. Server-side objectives (contribution quality, communication delay, and price) are encoded directly in the priority profile that the mechanism takes as input, so the TTC procedure structurally favors pairings that servers value highly. The theoretical guarantees of SCOPE-FL (Pareto efficiency and strategy proofness) are accordingly stated for the client side, while server-side and system-level outcomes, such as global model accuracy and convergence speed, are validated empirically in evaluation Section.

TTC is the unique mechanism in this setting known to simultaneously guarantee PE and SP while provide minimum instability (minimum number of blocking pairs among all SP and PE mechanisms) \cite{dougan2022robust}, ensuring that the system reaches its pareto efficient allocation while incentivizing all participants to report their preferences and capabilities truthfully. To accurately measure each participant's marginal contribution to global model, SCOPE-FL employs a scalable heuristic-based Shapley value method based on One-Round Reconstruction (OR) \cite{song2019profit}, customized for the HFL framework to enable fair and transparent reward distribution without the combinatorial complexity of exact Shapley computation. The entire assignment mechanism executes via blockchain smart contracts, providing the tamper-proof execution environment necessary for SP guarantees to be credible in practice and eliminating single points of failure inherent in centralized designs.

\begin{enumerate}

\item \textbf{Pareto efficient client selection.} We are the first to formulate HFL client selection as a two-sided school choice problem and solve it using TTC, transforming what was previously a stability-oriented, welfare-sacrificing process into one that guarantees a pareto efficient allocation for clients. This moves beyond unilateral, server-centric selection to a bilateral model where the preferences and priorities of both clients and servers drive the matching outcome.

\item \textbf{Simultaneous PE and SP.} SCOPE-FL is a pioneering HFL framework that simultaneously guarantees PE and SP with minimum instability. This unique dual guarantee ensures PE while making it a dominant strategy for all participants to report truthfully, so the welfare gains are realized in practice and not just in theory. 

\item \textbf{Scalable contribution evaluation.} To ensure equitable performance-based compensation, SCOPE-FL employs a scalable heuristic-based Shapley value method using OR, accurately measuring each client's marginal contribution to global model improvement. This contribution-proportional compensation ensures that clients who improve the global model most are rewarded most, creating a self-reinforcing alignment between individual participation incentives and system-wide welfare maximization.

\item \textbf{Decentralized tamper-proof execution.} The entire matching and reward mechanism is implemented via blockchain smart contracts, eliminating centralization risks and ensuring transparent, manipulation-resistant operation. We further identify storage write complexity, quantified by SSTORE operations on the EVM as the critical efficiency metric for matching algorithms on blockchain, a contribution that has been systematically overlooked in prior work \cite{irving1987efficient}.

\item \begin{sloppypar} \textbf{Comprehensive empirical validation.} We conduct extensive evaluation on MNIST, Fashion-MNIST, and CIFAR-10, demonstrating that SCOPE-FL outperforms DA, IAS, and other baseline methods across model accuracy, convergence rate, reward efficiency, gas consumption, and communication latency.\end{sloppypar}

\end{enumerate}

The paper’s following sections are structured as follows:                                                           
Section ~\ref{Related Work} provides a deep dive into related work. Section ~\ref{Background} provides the foundational background necessary to contextualize this study, while Section \ref{System Model} demonstrates the adopted system model. Section ~\ref{Evaluation} covers the experimental setup and findings. Finally, Section ~\ref{Conc} discusses the conclusion and future research directions.

\section{Related Work} \label{Related Work}
In the following, we survey related work on client selection and incentive mechanisms in FL systems, multi-server FL and HFL frameworks, blockchain-enabled FL systems, and clients' contribution evaluation. 

\subsection{Client Selection and Incentive Mechanisms in FL Systems}

Client selection has been widely recognized as a key determinant of system welfare in FL, yet existing mechanisms address welfare only partially. Optimization-based, importance-driven, clustering, and reinforcement learning approaches primarily maximize server-side utility while treating selection as a unilateral decision, ignoring client preferences and bilateral welfare \cite{li2024comprehensive, shi2023fairness, marnissi2024client, albelaihi2023deep}. Game-theoretic approaches move closer to bilateral modeling, for instance, Yellampalli et al. propose MAAIM, a DA-based incentive mechanism achieving stable client-server pairings through learning quality estimation, excelling at stability but not aiming to achieve PE. \cite{yellampalli2024client}. Similarly, Wehbi et al. employ matching game theory for bilateral selection based on accuracy and rewards, and later extend this to a mutual trust-based framework that significantly reduces untrusted client participation, but without PE or SP guarantees \cite{wehbi2022towards, wehbi2023towards}. Qu et al. propose COCS, a contextual combinatorial multi-armed bandit policy for HFL-specific client selection that handles uncertain network conditions and budget constraints, excelling at maximizing participating client count but not aiming to achieve bilateral preference satisfaction or pareto efficient matching \cite{qu2022context}. For contribution evaluation and reward distribution, Zahra et al. introduce Block-RACS, a blockchain-driven mechanism using influence function-based Shapley estimation that achieves computational efficiency but does not capture coalition effects essential for accurate marginal contribution assessment \cite{batool2023block}.

Beyond selection mechanics, the role of incentives has been analyzed broadly. Nair et al. survey auction, contract, and game-theoretic incentive mechanisms, emphasizing contribution evaluation and manipulation resistance \cite{nair2025incentivized}. Donahue et al. study optimality and stability in FL, identifying a fundamental tension: mechanisms that achieve equilibrium stability systematically sacrifice efficiency, leaving potential welfare gains unrealized \cite{donahue2021optimality}. DualGFL combines hedonic coalition formation with multi-attribute auction mechanisms to achieve Pareto-optimal coalition structures, demonstrating the potential of hybrid cooperative-competitive modeling, but without addressing SP at the client-server matching level \cite{chen2025dualgfl}. Diamanti et al. frame multi-server resource allocation as a Fisher market equilibrium problem guaranteeing proportional fairness and envy-freeness, excelling at resource pricing but not aiming to provide matching-theoretic welfare guarantees \cite{diamanti2025resource}.

Collectively, these works demonstrate the importance of incentive design in FL but leave the core welfare problem unresolved: no existing mechanism simultaneously achieves PE, SP, and accurate coalition-aware contribution evaluation in HFL.

\subsection{Multi-Server and Hierarchical Federated Learning}
The limitations of single-server FL, such as communication bottlenecks, vulnerability to failure, and unilateral client selection have motivated extensive research into multi-server and hierarchical architectures. Tang et al. propose a multi-UAV-assisted HFL framework combining Lyapunov optimization with deep reinforcement learning for joint client selection and server assignment, achieving significant reductions in training latency and energy consumption but without bilateral incentive compatibility \cite{tang2023multi}. HaghighiFard et al. extend this to vehicular networks with a multi-tiered architecture demonstrating improved scalability and non-IID handling, while Zuo et al.'s Spyker eliminates synchronization bottlenecks through fully asynchronous multi-server FL, reducing convergence time by up to 61\% in geo-distributed settings \cite{haghighifard2025hierarchical, zuo2024spyker}. Yang et al. propose HierMo, a three-tier momentum-accelerated HFL algorithm with a proven O(1/T) convergence rate that reduces total training time by 5–73\% over two-tier baselines, excelling at convergence efficiency but not aiming to address client selection incentives or matching-theoretic welfare guarantees \cite{yang2023hierarchical}. Hudson et al.'s Flight framework enables truly hierarchical multi-tier FL with over 60\% reduction in communication overhead and support for up to 2,048 devices, demonstrating impressive structural scalability without addressing the strategic behavior of participants in client selection \cite{hudson2025flight}.

These frameworks collectively advance HFL's structural capabilities, like convergence speed, communication efficiency, and fault tolerance but consistently leave the matching-level welfare problem unaddressed. Client selection remains either unilateral, stability-focused, or lacking SP, preventing the system from reaching its pareto efficient allocation.

\subsection{Blockchain-Enabled and Decentralized FL}

Blockchain has been extensively explored as infrastructure for trust, transparency, and decentralization in FL, motivated precisely by the welfare-destroying consequences of centralized execution identified in Section~\ref{intro}. Cai et al. survey blockchain-empowered FL, highlighting its resistance to single points of failure and ability to secure incentive mechanisms, while noting efficiency and storage challenges that motivate further optimization \cite{cai2025blockchain}. Zhang et al. categorize decentralized FL frameworks by consensus model and application domain, and Orabi et al. further emphasize blockchain's role in mitigating poisoning attacks and ensuring data integrity \cite{zhang2024decentralized, orabi2025adapting}. Wu and Seneviratne propose a scalable blockchain-based FL framework with automated registration, validation, and reward distribution for large-scale tasks \cite{wu2025blockchain}. Kasyap et al.'s PoIS consensus mechanism uses Shapley value-based model interpretation to evaluate contributions and detect malicious participants, restricting attack success rates below 5\% even under 90\% adversarial participation \cite{kasyap2022efficient}. Cam and Kiet's FlwrBC and Wang et al.'s consortium blockchain mechanism further demonstrate blockchain's utility for transparent incentive enforcement in general and healthcare FL settings respectively \cite{cam2023flwrbc, wang2025multi}.

Despite these advances, existing blockchain-based FL solutions focus primarily on trust and security rather than on enforcing matching-theoretic welfare guarantees. So, even theoretically sound mechanisms are undermined by centralized execution that makes their guarantees
unenforceable in practice. Moreover, the on-chain storage complexity of matching algorithms has not been systematically analyzed, leaving an open question about which mechanisms are most suitable for blockchain deployment (a gap that SCOPE-FL directly addresses through its SSTORE-based complexity analysis).

\subsection{Clients’ Contribution Evaluation}
Accurate contribution evaluation is a prerequisite for welfare-maximizing reward distribution in HFL since without it, compensation is decoupled from actual marginal impact, destroying the incentive compatibility that PE and SP are designed to preserve. Block-RACS introduces an influence function-based mechanism that is computationally efficient and integrates reputation tracking, but deviates from true contribution values in deep non-convex models and neglects coalition effects among clients \cite{batool2023block}. MAAIM relies on learning quality estimation based on loss reduction, which is lightweight and effective for client ranking but captures only individual performance in isolation, missing the coalition-level interactions essential for Shapley-inspired contribution evaluation \cite{yellampalli2024client}. 

Multi-round reconstruction approximates Shapley values more accurately by considering long-term participation, but incurs prohibitive computational and memory costs that limit scalability in large client populations \cite{song2019profit}. The OR method best balances accuracy, coalition awareness, and per-round efficiency among existing approximations, but its $2^n$ subset evaluation complexity remains intractable at scale when applied over an entire client population \cite{song2019profit}. Therefore, no contribution evaluation method achieves accuracy, scalability, and coalition awareness together.

Together, mentioned three gaps define the precise problem space that SCOPE-FL occupies, one that no prior work has addressed in its entirety.

\section{Background} \label{Background}
In the following section we discuss the school choice problem and provide details about the well-known TTC algorithm with an example.
\subsection{School Choice Problem}

The school choice problem represents a fundamental assignment challenge within the broader framework of assignment theory. This problem involves two distinct sets: a collection of individuals (students) and a corresponding set of objects (schools). Each individual possesses a preference ordering over the available objects, while each object has an associated capacity constraint and maintains priorities over the individuals based on specific criteria \cite{haeringer2018market}.

The formal structure of a school choice problem can be represented as the tuple $(I, S, q, P, \pi)$, where:
\begin{itemize}
    \item \textbf{I} = $\{i_1, i_2, ..., i_n\}$: A finite set of students (individuals)
    \item \textbf{S} = $\{s_1, s_2, ..., s_m\}$: A finite set of schools
    \item \textbf{q} = $\{q_{s_1}, q_{s_2}, ..., q_{s_n}\}$: A capacity vector specifying the maximum number of students each school can accommodate 
    \item \textbf{P} = $\{P_{i_1}, P_{i_2}, ..., P_{i_n}\}$: A preference profile representing the strict preferences of students over schools
    \item $\boldsymbol{\pi}$ = $\{\pi_{s_1}, \pi_{s_2}, ..., \pi_{s_m}\}$: A priority structure indicating each school's ordering over students
\end{itemize}

For any student $i$, the preference relation $P_i$ is defined as a strict ordering over the set $S \cup \{i\}$. The inclusion of the student itself in their preference ordering allows for the possibility of self-matching, representing scenarios where a student might prefer remaining unmatched rather than being assigned to certain schools they consider undesirable \cite{haeringer2018market}.

An assignment is formally represented as a mapping $\mu: I \cup S \rightarrow 2^{I \cup S}$ satisfying the following conditions for each $i \in I$ and $s \in S$:
\begin{enumerate}
\item $\mu(i) \in S \cup \{i\}$ and $\mu(s) \in 2^I$

\item $\mu(i) = s$ if and only if $i \in \mu(s)$

\item $|\mu(s)| \leq q_s$
\end{enumerate}

Two fundamental properties characterize desirable assignment mechanisms: efficiency and stability. An assignment $\mu$ is considered efficient if no alternative assignment $\mu'$ exists where all students weakly prefer $\mu'$ over $\mu$ (meaning students either view both assignments as equally preferable or strictly prefer $\mu'$) and at least one student strictly prefers $\mu'$ over $\mu$. This ensures that no Pareto-improving reassignment is possible \cite{haeringer2018market}.

An assignment $\mu$ is deemed stable if it simultaneously satisfies three conditions:

\begin{enumerate}
\item Individual Rationality: For every student $i \in I$, their assignment $\mu(i)$ is at least as preferred as remaining unmatched.
\item Non-wastefulness: If a student prefers a particular school to their current assignment, then that school must have reached its capacity.
\item Absence of Justified Envy: If student $i$ prefers school $s$ over her current assignment, then all students assigned to school $s$ must have higher priority than student $i$ according to $\pi_s$.
\end{enumerate}

A fundamental impossibility result demonstrates that no assignment mechanism can simultaneously guarantee both efficiency and stability \cite{abdulkadirouglu2003school,bichler2017market, haeringer2018market}. This impossibility forces practical mechanisms to favor one property over the other, so the choice depends on which better serves the application at hand.

\subsection{Top Trading Cycle Mechanism}

One prominent approach is the TTC mechanism, which is known to produce pareto efficient allocations, though it does not always satisfy stability \cite{abdulkadirouglu2003school}. By contrast, DA guarantees stable outcomes but, as noted earlier, may fail to achieve efficiency. For the client selection problem in federated learning, we favor efficiency over stability and therefore adopt TTC since it optimizes clients' welfare in the Pareto sense and guarantees Pareto efficient outcome. Operationally, TTC is represented as a directed graph whose nodes are students and schools: at the start of each iteration, every student points to their most preferred school among those still available, and each school points to the highest-priority student among those not yet assigned. It can be shown that at least one directed cycle must exist. We select any cycle, and for every student in that cycle, assign them to the school they point to; all students in the cycle are then removed from the graph. For each school in the cycle, its capacity is reduced by one, and if its capacity reaches zero the school node is removed. The procedure then repeats with the remaining students and schools. Students re-point to their top remaining choice, schools re-point to their highest-priority remaining student, and cycles are cleared in the same way. The mechanism terminates when every student is assigned or when no school seats remain \cite{haeringer2018market}. A step-by-step illustration of TTC is provided in Appendix A with an example.

\section{System Model} \label{System Model}
This section begins with outlining the problem scenario. Next, we describe the models for both servers and clients, followed by the formulation of the problem. To help readers follow along, Table \ref{tab:notations} lists all the symbols and notation used throughout the model.

\begin{table}[h]
% [width=.9\linewidth,cols=2,pos=h]
\caption{Summary of notations}
\label{tab:notations}
\begin{tabularx}{\columnwidth}{lX}
\toprule
\textbf{Parameters} & \textbf{Description} \\
\midrule
$C$, $S$                            & Sets of clients and tier-2 servers \\
$s_i$                               & A single tier-2 FL server \\
$c_j$                               & A single client device \\
${P^r}$, ${Q^r}$                            & Preference \& priority matrices in round r \\
$\mathit{Cap}_{s_i}$                & Capacity of tier-2 server $i$ \\
$r_n$                               & A federated server communication round \\
$R$                                 & Total number of FL rounds \\
$I_x$                               & Assigned weight to parameter $x$ \\
$A^r$                                 & Assignment mapping in round r \\
$C_{s_i}$                           & Assigned client set to tier-2 server $i$ \\
$G$                                 & Directed graph for TTC \\
$SC$                                & Smart contract \\
$\theta$                            & Model parameters \\
$\phi$                            & contribution index of each client\\
$\Delta$                            & Changes in model state \\
$M$                                 & Model state \\
$D$                                 & Dataset of each client \\
$n_{c_j}$                           & data size of client $j$\\
$n_{s_i}$                           & data size of server $i$\\
$\ell$                              & Loss function \\
$U(\cdot)$                          & Model utility function (e.g., validation accuracy) \\
$K$                                 & Shapley normalization constant \\
$p^{s_i}_{c_j}$                     & Offered price by server $i$ to clients \\
$p^{c_j}_{s_i}$                     & Requested price by client $j$ to contribute \\
$\mathit{Net\_delay}^{s_i}_{c_j}$   & Average network delay between server $i$ and client $j$ \\
$\mathit{Obj}^{s_i}_{c_j}$                & Objective function of tier-2 server $i$ with respect to client ${j}$\\
$O_{c_j}^{s_i}$                           & Total normalized gain of client $j$ with respect to server $i$ \\
$E_{s_i}$                           & Overall objective function of each tier-2 server ${i}$ across all assigned  clients \\
$W^r(A)$                            & System welfare function at round $r$ \\
\bottomrule
\end{tabularx}
\end{table}

The multi-server FL frameworks employ multiple servers that interact with disjoint client subsets, reducing computational and communication overhead. Among these, HFL systems, such as HierFAVG adopt a layered architecture in which a principal server maintains the global model \cite{liu2020client}. In each round, tier-2 (edge) servers aggregate updates from their client groups and forward the results to the principal server, which performs a second-level aggregation to produce the new global model. Most existing HFL frameworks operate synchronously, requiring all servers to finish aggregating before the global update is finalized, though asynchronous variants have been proposed to handle stragglers. In geo-distributed settings, this design reduces wide-area communication, improves scalability, and mitigates client heterogeneity at the regional level \cite{zuo2024spyker}.

Within this architecture, both clients and servers hold incentive requirements. Servers prefer clients that demand lower rewards and offer higher marginal contributions to the global model, while clients prefer servers offering higher compensation and lower latency. The matching should therefore be contribution-aware, PE, and SP, and must execute in a tamper-proof environment, which we realize through blockchain smart contracts. We detail the server, client, and welfare models and the resulting matching formulation in the following subsections.

\subsection{Server Model in SCOPE-FL}

In our HFL framework, servers are organized into two layers: tier-2 servers and a top server. Each tier-2 server is responsible for aggregating local model updates from its assigned clients, evaluating their contributions using the OR-based mechanism \cite{song2019profit} over its local client set, and scoring them according to factors such as network delay, price, and contribution values. These aggregated updates and contribution scores are then transmitted to the top server. The top server serves as the principal coordinator and it collects the aggregated models from all tier-2 servers, performs a weighted global aggregation, and updates the global model accordingly. It also executes the smart contract–based assignment mechanism to ensure efficient client–server associations for subsequent rounds. Finally, the updated global model and assignment decisions are redistributed back to all servers and clients, enabling the next iteration of training.

In SCOPE-FL, the OR method is executed independently at each tier-2 server over only its assigned client set, not over the entire client population. So, while the exact Shapley value computation requires evaluating all $2^n$ subsets of clients, which is intractable for large n, our hierarchical architecture naturally bounds this complexity. 

The Algorithm \ref{alg:scalable_or} proceeds in three phases. In the first phase, each assigned client $j \in C_{s_i}$ trains a local model $M^{(r)}_j$ using its local dataset and computes a model update $\Delta^{(r)}_j = M^{(r)}_j - M^{(r)}$, which is sent to server $s_i$. In the second phase, server $s_i$ aggregates these updates to form the new global model $M^{(r+1)}$ and simultaneously reconstructs approximate coalition models $M^{(r+1)}_{Sub}$ for every subset $Sub \subseteq C_{s_i}$ using the stored gradients without requiring any additional training rounds. This one-round reconstruction is the key efficiency property of the OR method: coalition models are derived directly from existing updates rather than retraining from scratch. In the third phase, the server computes each client's contribution index $\phi_j$ using the Shapley formula, measuring the marginal utility gain of adding client $j$ to each possible coalition $Sub \subseteq C_{s_i} \setminus \{j\}$.
 
Since the subset enumeration in phase two is bounded by $2^{\mathit{Cap}_{s_i}}$ rather than $2^{|C|}$, the system-wide complexity reduces from $O(2^{|C|})$ globally to $O(|S| \cdot 2^{\mathit{Cap}_{\max}})$, where $|S|$ is the number of tier-2 servers and $\mathit{Cap}_{\max}$ is the maximum server capacity. Since server capacity is a bounded system parameter independent of total client population size, contribution evaluation remains tractable as the number of clients grows, making SCOPE-FL significantly more scalable than flat FL systems where OR would need to operate over the entire client pool.
 
\begin{algorithm}[t]
\caption{Scalable OR (One-Round Reconstruction) in SCOPE-FL}
\label{alg:scalable_or}
\begin{algorithmic}[1]
\REQUIRE $B$: local minibatch size, $E$: number of local epochs, $\eta$: learning rate
\STATE \textbf{Server $s_i$ executes:} \algorithmiccomment{Runs independently at each tier-2 server}
\STATE \textbf{ Calculate the Local Model for assigned clients only}
\STATE $C_{s_i} \leftarrow$ assigned client set of $s_i$, $|C_{s_i}| \leq \mathit{Cap}_{s_i}$
\STATE Initialize $M^{(0)}$, $\{M^{(0)}_{\mathit{Sub}} \mid \mathit{Sub} \subseteq C_{s_i}\}$
\FOR{each round $r \leftarrow 0, 1, 2, \ldots, R-1$}
    \STATE Send $M^{(r)}$ to all clients $j \in C_{s_i}$
    \STATE $M^{(r)}_j \leftarrow \textsc{Client~Update}(j, M^{(r)})$ for $j \in C_{s_i}$
    \STATE $\Delta^{(r)}_j \leftarrow M^{(r)}_j - M^{(r)}$ for $j \in C_{s_i}$
    \STATE $M^{(r+1)} \leftarrow M^{(r)} + \sum_{j \in C_{s_i}} \frac{|D_j|}{\sum_{k \in C_{s_i}} |D_k|} \cdot \Delta^{(r)}_j$
    \FOR{each subset $\mathit{Sub} \subseteq C_{s_i}$}
        \STATE $\Delta^{(r)}_{\mathit{Sub}} \leftarrow \sum_{j \in \mathit{Sub}} \frac{|D_j|}{\sum_{k \in \mathit{Sub}} |D_k|} \cdot \Delta^{(r)}_j$
        \STATE $M^{(r+1)}_{\mathit{Sub}} \leftarrow M^{(r)}_{\mathit{Sub}} + \Delta^{(r)}_{\mathit{Sub}}$
    \ENDFOR
\ENDFOR
\STATE \textit{/* Calculate the CIs */}
\FOR{$j \leftarrow 1, 2, \ldots, |C_{s_i}|$}
    \STATE $\phi_j = K \cdot \sum_{\mathit{Sub} \subseteq C_{s_i} \setminus \{j\}} \frac{U(M^{(R)}_{\mathit{Sub} \cup \{j\}}) - U(M^{(R)}_{\mathit{Sub}})}{\binom{|C_{s_i}|-1}{|\mathit{Sub}|}}$
\ENDFOR
\STATE \textbf{Return} $M^{(r)}$ and $\phi_1, \phi_2, \ldots, \phi_{|C_{s_i}|}$
\STATE \textbf{ClientUpdate}$(j, M)$\textbf{:}
\STATE $\mathcal{B} \leftarrow$ (split $D_j$ into batches of size $B$)
\FOR{each local epoch $e \leftarrow 1, 2, \ldots, E$}
    \FOR{batch $b \in \mathcal{B}$}
        \STATE $M \leftarrow M - \eta \nabla \ell(M;\, b)$
    \ENDFOR
\ENDFOR
\STATE \textbf{Return} $M$ to server
\end{algorithmic}
\end{algorithm}

As previously mentioned, federated tier-2 servers wish to maximize model accuracy while paying the least possible reward to client devices and minimize overall network delay, at the same time. Note that network delay between clients and servers can be calculated by sending a short message and checking the Round-Trip Time (RTT) of each packet.  Therefore, the server $s_i$ is interested in maximizing the following objective function $Obj_{c_j}^{s_i}$ for each assigned client during each FL round $r_n$ in Equation (\ref{eq:4}). 

Communication overhead in FL systems is influenced by multiple factors, including model size, bandwidth availability, network topology, and convergence rate \cite{ayeelyan2024federated, zuo2024spyker}. Among these, transmission latency between clients and servers (captured by round-trip time RTT) is one of the most directly controllable and impactful factors, as it determines how long each round of model upload and download takes regardless of other system parameters. In SCOPE-FL, this is captured through the $\mathit{Net\_delay}^{s_i}_{c_j}$ term explicitly present in both the server objective~(\ref{eq:4}) and the client objective~(\ref{eq:11}). By minimizing $\mathit{Net\_delay}$ in the matching objective, SCOPE-FL's TTC mechanism implicitly tries to minimize communication overhead at the assignment level, preferring client-server pairings with lower RTT-based transmission costs. Communication overhead is therefore a first-class component of both utility functions and is formally integrated into the welfare objective $W^r(A)$ defined in Section~\ref{sec:welfare}.

It should be noted that since engaged parameters in formulas are measured in different units, each part is normalized between zero and one with the help of min-max normalization. The general formula for performing min-max normalization is as Equation (\ref{eq:8}):

\begin{equation}
\label{eq:8}
x' = \frac{x - \min(x)}{\max(x) - \min(x)}
\end{equation}

The weights $I_x$, $I_{x'}$, $I_{x''}$ in Equation (\ref{eq:4}) are configurable parameters 
that allow each server to independently express its operational priorities. 
In real-world deployments, servers may assign different weights based on their 
specific requirements. servers may prioritize contribution 
quality ($I_x \gg I_{x'}, I_{x''}$), cost-conscious providers may emphasize reward 
minimization ($I_{x''} \gg I_x, I_{x'}$), and edge servers may focus on latency 
($I_{x'} \gg I_x, I_{x''}$). For experimental consistency, we set uniform weights ($I_x = I_{x'} = I_{x''} = 1/3$) across all servers to provide an unbiased baseline for mechanism comparison.

\begin{equation}
\label{eq:4}
Obj_{c_j \in C_{s_i}} ^{s_i} = I_x \phi_j - I_{x'} Net\_delay_{c_j}^{s_i} - I_{x''} p_{s_i}^{c_j}
\end{equation}

Therefore, our solution should maximize Equation (\ref{eq:6}), for each tier-2 server $s_i \in S$ across its assigned clients. As a final note, Constraint (\ref{eq:7}) limits each tier-2 server to select at most $Cap_{s_i}$ clients per round.

\begin{equation}
\label{eq:6}
E_{s_i} = \sum_{c_j \in C_{s_i}}^{N_{sel}} Obj_{c_j}^{s_i}
\end{equation}

\begin{equation}
\label{eq:7}
C_{s_i} \leq Cap_{s_i}
\end{equation}

Additionally, to avoid further complication and overhead we use the FedAvg formula to aggregate client weights in tier-2 servers and tier-2 servers' weights in the top server, according to Equations (\ref{eq:fedavg_tier2}) and (\ref{eq:fedavg_top}). FedAvg is the baseline federated learning algorithm where, in each round, the server sends the current global model to a subset of clients, each client trains it locally for a few epochs on their own data, then sends the updated model back. The server combines these client models by averaging them (giving more weight to clients with more data) to produce a new global model, and repeats this process until convergence \cite{jung2025federated}. Consider that tier-2 server $s_i$ has client set $C_{s_i}$, client $c_j$ has data size $n_{c_j}$, and server $s_i$ total available data is $n_{s_i} = \sum_{c_j \in C_{s_i}} n_{c_j}$. The total data across all clients is $N = \sum_{s_i \in S} n_{s_i}$. We define $\Delta_{c_j}^{(r)} = M_{c_j}^{(r+1)} - M^{(r)}$ as the difference between client $c_j$'s local model and the current global model after training, and $\Delta_{s_i}^{(r)} = M_{s_i}^{(r+1)} - M^{(r)}$ as the difference between the global model and tier-2 server $s_i$'s local model after aggregation.

\begin{equation}
\label{eq:fedavg_tier2}
M_{s_i}^{(r+1)} = M^{(r)} + \sum_{c_j \in C_{s_i}} \frac{n_{c_j}}{n_{s_i}} \Delta_{c_j}^{(r)}
\end{equation}

\begin{equation}
\label{eq:fedavg_top}
M^{(r+1)} = M^{(r)} + \sum_{s_i \in S} \frac{n_{s_i}}{N} \Delta_{s_i}^{(r)}
\end{equation}

In comparison to FedAvg, we have FedProx and FedRelax, two algorithms that are developed for better dealing with data heterogeneity. The core update formula for FedProx is demonstrated in Equation (\ref{eq:10}). FedProx is an iterative federated learning algorithm that alternates between the separate training of local models, followed by combining the updated local model parameters. It was designed to address a central challenge in FedAvg: selecting an appropriate number of local updates. FedProx uses a proximal operator for training instead of standard Stochastic Gradient Descent (SGD), making it particularly effective for FL applications with high levels of heterogeneity among devices' computational capabilities and statistical properties of their local datasets \cite{jung2025federated}.

\begin{equation}
\label{eq:10}
M_j^{(r+1)} = \arg\min \left[ \ell(M; D_j) + \frac{1}{\eta} \|M - M^r\|_2^2 \right]
\end{equation}

In the Equation (\ref{eq:10}), $\frac{1}{\eta} \|M - M^r\|_2^2$ is the proximal regularization term and $\ell(M; D_j)$ is the loss function evaluated on client $j$'s local dataset $D_j$. On the other hand, FedRelax applies block-coordinate minimization to solve the Generalized Total Variation Minimization (GTVMin) problem. It is designed to handle both parametric and non-parametric models, making it model-agnostic. The algorithm exploits the structure of the objective function by updating local model parameters in parallel while keeping the parameters of the neighbors fixed \cite{jung2025federated}.

\subsection{Client Model in SCOPE-FL}

The main objective of client devices is to maximize their earned reward alongside reducing network latency. In (\ref{eq:11}), for each client $c_j$ and its matched server $s_i$, we demonstrate the objective function of each client. Note that client devices aim to maximize the objective function in Equation (\ref{eq:11}), in which the values of $I_x$, $I_{x'}$ will be adjusted according to each client's preferences. For example, if client $c_j$ prefers to maximize the reward, $I_x$ should be much greater than $I_{x'}$ ($I_x \gg I_{x'}$), and vice versa. In this scenario, we assign (1/2) to each parameter.

\begin{equation}
\label{eq:11}
O_{c_j}^{s_i} = I_x p_{c_j}^{s_i} - I_{x'} Net\_delay_{c_j}^{s_i}
\end{equation}

In such a situation, each client $j$ can be assigned to at most one federated server, per each $r_n$ to satisfy (\ref{eq:12}).

\begin{equation}
\label{eq:12}
|A^{r_n}(c_j)| \leq 1 \quad \forall c_j \in C, \, \forall r_n
\end{equation}

It should be noted that each client device runs a simple Convolutional Neural Network (CNN) model that will be described in detail in the evaluation, and the client model above assumes fully rational utility-maximizing agents, meaning each client is modeled as acting to maximize $O_{c_j}^{s_i}$ without error.

\subsection{Welfare Formulation in SCOPE-FL}
\label{sec:welfare}
 
Beyond individual utility maximization, The system-level goal of SCOPE-FL is to reach allocations from which no Pareto improvement is possible during each federated learning round. We define the global welfare function as the sum of realized utilities of all matched participants, subject to capacity and assignment feasibility constraints.

The system welfare at round $r$ is defined as:
\begin{equation}
W^r(A) = \sum_{s_i \in S} \sum_{c_j \in C_{s_i}} (Obj_{c_j}^{s_i}
        + \ O_{c_j}^{s_i})
\label{eq:welfare}
\end{equation}
$Obj_{c_j} ^{s_i}$ is the server objective function defined in~(\ref{eq:4}), and $O_{c_j}^ {s_i}$ is the client objective function defined in~(\ref{eq:11}), subject to:
\begin{enumerate}
    \item $|C_{s_i}| \leq \mathit{Cap}_{s_i}$ for all $s_i \in S$ \hfill (capacity feasibility)
    \item Each client is assigned to at most one server \hfill (unique assignment)
\end{enumerate}

The welfare function $W^r(A)$ captures the bilateral nature of SCOPE-FL's matching problem: server utilities reflect model quality, communication cost, and reward efficiency, while client utilities reflect compensation and latency preferences. An assignment that maximizes $W^r(A)$ in the Pareto sense ensures that no reallocation can improve any participant's utility without reducing another's.

As formalized in Section~\ref{4.4} and proven in Theorem 1, in section \ref{thm:1}, SCOPE-FL, solves this problem by formulating client selection as a school choice problem and through the TTC mechanism.

It is important to clarify the precise relationship between the TTC mechanism and the welfare function $W_r(A)$. SCOPE-FL does not optimize $W_r(A)$ directly; rather, it guarantees Pareto efficiency with respect to the client preference profile $P^r$, which is constructed as a monotone transformation of the client utilities $O^{s_i}_{c_j}$. By Theorem 1, the resulting assignment $A^{r+1}$ admits no reallocation that makes any client better off without making another client worse off under the submitted profiles. Accordingly, we employ $W_r(A)$ not as the objective the mechanism provably optimizes, but as an empirical aggregate that quantifies the realized utility of assignments, and we report its constituent components (reward and latency) as indicators of realized welfare in our evaluation section.

This welfare guarantee is only meaningful under truthful preference revelation. As established in Theorem 2, SP ensures that participants have no incentive to misrepresent $P^r$ or $Q^r$, which is the necessary condition for the pareto efficient properties of TTC to hold in practice. Without SP, participants could corrupt the preference profile fed to TTC, producing an assignment that is pareto efficient only with respect to false inputs and therefore welfare-destroying in reality.

A natural question is what protects server interests, given that the formal guarantees of Theorems~1 and~2 concern only the client side. In SCOPE-FL, servers are not strategic agents but objects endowed with capacities and priorities \cite{abdulkadirouglu2003school}; their interests therefore enter the mechanism through the priority profile $Q^r$ rather than through the efficiency guarantee itself. Since $Q^r$ ranks clients by the very quantities servers care about (contribution scores, network delay, and requested price) and since a TTC cycle forms only when a server's highest-priority remaining client points back to it, the mechanism structurally channels high-contribution, low-cost clients toward the servers that value them most. Whether this translates into strong system-level outcomes is an empirical question, which evaluation section answers affirmatively; client-side Pareto efficiency coincides with the fastest convergence and highest final accuracy among all evaluated mechanisms.

\subsection{Problem Formulation as a School Choice} \label{4.4}
In this section, we formally map the client-server assignment challenge in hierarchical multi-server federated learning to the school choice framework, enabling us to leverage the TTC mechanism for pareto efficient and strategy-proof assignments. The overall architecture of SCOPE-FL has been illustrated in Figure \ref{Fig:SCOPE-FL Architecture}. This figure illustrates a HFL system where clients are matched to servers through a blockchain-based TTC mechanism. 

The architecture consists of three tiers: clients at the bottom level who participate in local training and price negotiations, multiple Tier-2 servers in the middle that aggregate client contributions and they are distributed in various geographical locations, and a top server that coordinates the global model. Clients under the coverage of each Tier-2 server send their model updates to their assigned Tier-2 servers, which then forward aggregated models, contribution scores, and clients' scores to the top server. The top server maintains the global model and communicates with a blockchain network running a TTC smart contract that processes clients' preferences for servers and servers' priorities over clients to compute fair assignments for the next round. 

\begin{figure*}[t]  % t=top of page, b=bottom of page
    \centering
    \includegraphics[width=1\textwidth]{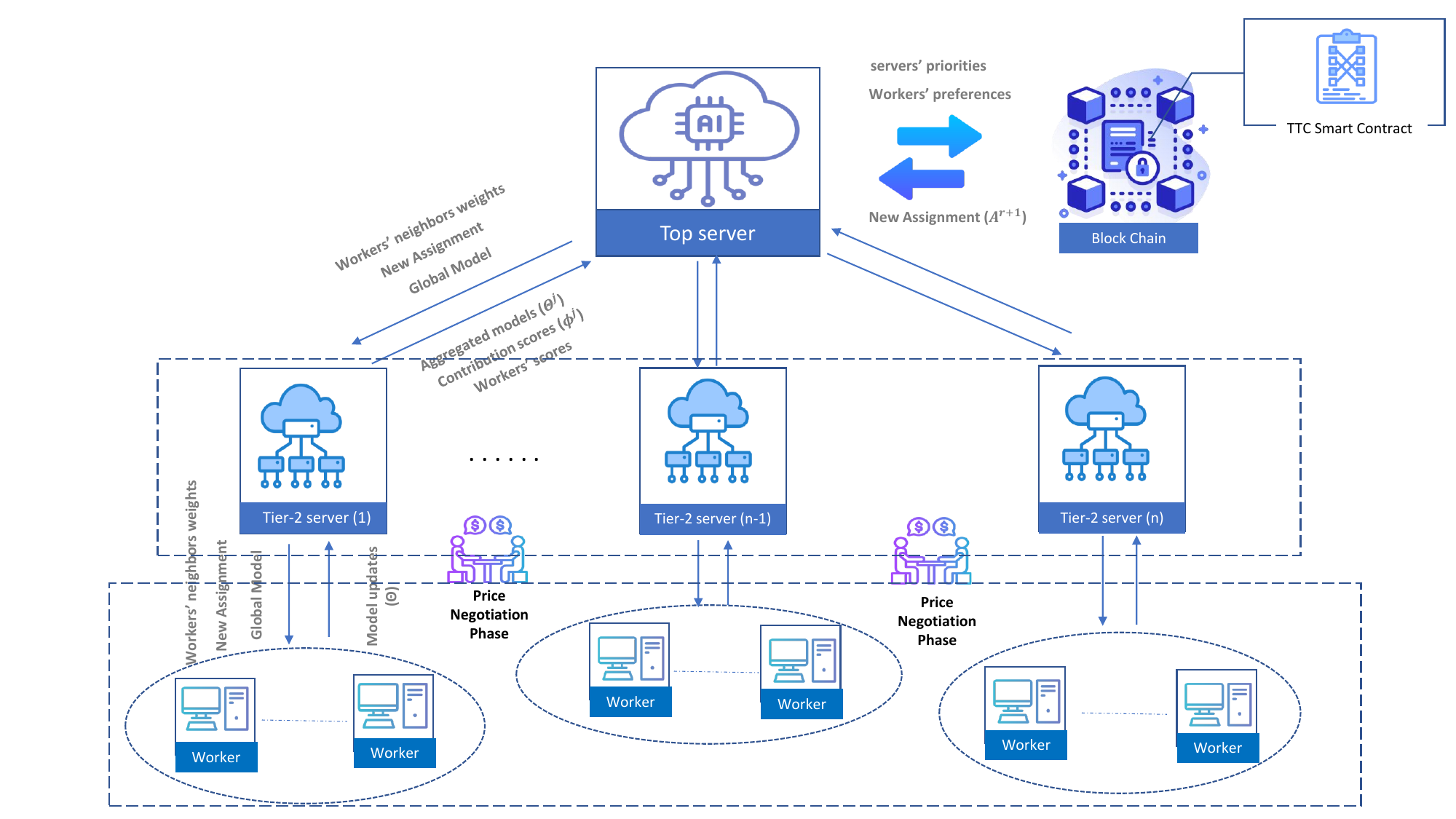}
    \caption{SCOPE-FL Architecture}
    \label{Fig:SCOPE-FL Architecture}
\end{figure*}

\subsubsection{Mapping FL Components to School Choice Elements}
We establish the following correspondence between federated learning entities and school choice problem components. Algorithm \ref{alg:scope-fl} illustrates the whole procedure of SCOPE-FL.

\textbf{1) Students $\leftrightarrow$ Client Devices (Workers):} The set of client devices C = $\{c_1, c_2, \ldots, c_n\}$ corresponds to the set of students seeking assignment. Each client $c_j$ possesses computational resources, local data $D_j$, and preferences over available servers.

\textbf{2) Schools $\leftrightarrow$ Tier-2 Servers:} The set of tier-2 federated learning servers S = $\{s_1, s_2, \ldots, s_m\}$ corresponds to the set of schools. Each server $s_i$ has a limited capacity ${Cap}_{s_i}$ representing the maximum number of clients it can accommodate in a given training round, analogous to school enrollment quotas.

\textbf{3) Student Preferences $\leftrightarrow$ client Preferences:} Each client $c_i$ maintains a strict preference ordering $P_{c_j}$ over the available servers. As described in Algorithm \ref{alg:scope-fl}, these preferences are constructed by ranking servers according to a utility function that combines normalized network delay and normalized price where higher values indicate more preferred servers. This preference reflects client's dual objectives: minimizing communication latency and maximizing monetary rewards.

\textbf{4) School Priorities $\leftrightarrow$ Server Priorities:} Each server $s_i$ maintains a priority ordering $Q_{s_i}$ over clients, constructed by ranking clients according to three normalized criteria (Algorithm \ref{alg:scope-fl}). Here, $\phi_j^{r-1}$ represents the OR-based contribution score from the previous round (or initialized uniformly in round 1). Unlike classical school choice where priorities are administrative rankings, our server priorities capture valuation based on efficiency, cost, and contribution quality.

\textbf{5) Capacity Constraints:} Each server $s_i$ has a maximum capacity ${Cap}_{s_i}$, enforced by the smart contract (Algorithm \ref{alg:TTC Optimal Assignment Smart Contract}). When a server reaches capacity through TTC assignments, it is removed from the graph, preventing over-subscription.

\textbf{6) Assignment Outcome:} The TTC mechanism produces an assignment mapping A$^{r+1}$ where A$^{r+1}$[$c_j$] = $s_i$ indicates that client $c_j$ is assigned to server $s_i$ for round r+1. This assignment respects capacity constraints and satisfies PE and SP properties.

\subsubsection{Dynamic Preference Updates}
One of the most important aspects of our approach is the dynamic evolution of preferences and priorities across training rounds. As shown in Algorithm \ref{alg:scope-fl}:

\textbf{a) Contribution-Based Evolution:} client contribution scores $\phi_j^r$ are recalculated each round using the OR method, influencing server priorities in subsequent rounds. High-contributing clients become more attractive to servers over time.

\textbf{b) Price Negotiation Integration:} The agreed price between client $c_j$ and server $s_i$ is contribution-weighted, creating a dynamic pricing mechanism where the base negotiated price (average of client's request and server's proposal) is scaled by a contribution-aware multiplier centered at 1. Specifically, a client whose contribution score $\phi_j^{r-1}$ equals the round mean $\bar{\phi}$ receives exactly the midpoint price, while above-average contributors receive a premium and below-average contributors receive proportionally less. This ensures higher-contributing clients receive better compensation, naturally incentivizing quality participation while maintaining fairness in the reward distribution.

\textbf{c) Adaptive System}: Our formulation represents a repeated game where assignments, contributions, and preferences co-evolve. The TTC mechanism is re-executed each round with updated profiles, ensuring continuous adaptation to network conditions and contribution dynamics.

\subsubsection{TTC Smart Contract}
The school choice problem formulated above is solved by Algorithm \ref{alg:TTC Optimal Assignment Smart Contract}, which implements the TTC mechanism on blockchain via smart contracts. The algorithm receives P$^r$, Q$^r$, and C as inputs, constructs a directed graph G where clients point to their most preferred available servers and servers point to their highest-priority available clients, and iteratively identifies and clears cycles until convergence. The resulting assignment A$^{r+1}$ guarantees PE, SP, and the minimum instability as demonstrated in the next section.

All matching mechanisms were implemented using smart contracts written in Solidity which are EVM-compatible. This blockchain-agnostic design enables deployment flexibility, allowing the most suitable network to be selected based on transaction costs, throughput, security, and application-specific requirements.

Three deployment scenarios are considered. First, Polygon that is a layer-2 scaling solution for Ethereum that processes transactions on a separate blockchain before settling them on the Ethereum mainnet. It offers faster transaction speeds (up to 65,000 transactions per second) and significantly lower fees compared to Ethereum's main network. \cite{werth2023review}. Second, Improved TrustChain that is considered for resource-constrained environments. Its DAG-based structure and local-chain-only design significantly reduce computational overhead and energy consumption, while the integrated NetFlow accounting mechanism provides resilience against both Sybil and whitewashing attacks without relying on traditional consensus protocols \cite{ghazi2022improved, ghazi2022suitability}. 

\begin{algorithm}[H]
\caption{SCOPE-FL with OR Method}
\label{alg:scope-fl}
\begin{algorithmic}[1]
\STATE Deploy $TTC\_SmartContract(Cap) \rightarrow SC$
\STATE Initialize: $TopServer$, $S$, $C$
\STATE Initialize: $\phi_j^0 \leftarrow 1$ for all $c_j \in C$ \& $r_n \leftarrow 1$
\WHILE{not converged}
    \STATE \textbf{PHASE A: Distributed Training \& OR Assessment}

    \FORALL{client $c_j \in C$ in parallel}
        \STATE $\theta^{r_n}_j \leftarrow TrainLocalModel(\theta_{global}, D_j)$
        \STATE \textbf{Price Negotiation:}
        \STATE Client $c_j$ requests price $p^{c_j}_{s_i}$
        \STATE Server $s_i$ proposes price $p^{s_i}_{c_j}$
        \IF{$c_j$ assigned to $s_i$ AND $|C_{s_i}| \leq Cap_{s_i}$}
            \STATE agreed price $\leftarrow \left\lfloor\left(1 + \frac{\phi_j^{r_n - 1} - \bar{\phi}}{\bar{\phi}}\right) \cdot \frac{p^{c_j}_{s_i} + p^{s_i}_{c_j}}{2}\right\rfloor$
            \STATE Accept negotiation
        \ELSE
            \STATE Reject negotiation
        \ENDIF
        \STATE Send $ModelUpdate(\theta^{r_n}_j, \ell_j, acc_j)$ to assigned server $s_i$
    \ENDFOR

    \FORALL{server $s_i \in S$ in parallel}
        \STATE $\{\theta^{r_n}_j\} \leftarrow$ Collect client updates from $C_{s_i}$
        \STATE $\theta^{r_n}_{s_i} \leftarrow Aggregate~CNN(\{\theta^{r_n}_j\})$
        \STATE $\phi^{r_n}_j \leftarrow Calculate ~OR ~Contribution(\{\theta^{r_n}_j\}, \theta_{global})$ \quad $\forall\, c_j \in C_{s_i}$
    \ENDFOR

    \STATE \textbf{PHASE B: Global Coordination \& Assignment}
    \STATE Collect all $(\theta^{r_n}_{s_i},\, \phi^{r_n}_j,\, score_{s_i})$ at $TopServer$
    \STATE $\theta_{global} \leftarrow WeightedGlobalAggregate(\{\theta^{r_n}_{s_i}\},\, \{\phi^{r_n}_j\})$

    \STATE \textbf{Build Client Preferences:}
    \FORALL{client $c_j \in C$}
        \STATE $P^{r_n}_{c_j} \leftarrow$ Rank servers by: \\
        \quad \begin{tabular}[t]{@{}l@{}}
            $(1 - \text{normalized } Net\_delay^{s_i}_{c_j})$ \\
            ${}+ (1 - \text{normalized } p^{s_i}_{c_j})$
        \end{tabular}
    \ENDFOR

    \STATE \textbf{Build Server Priorities:}
    \FORALL{server $s_i \in S$}
        \STATE $Q^{r_n}_{s_i} \leftarrow$ Rank clients by: \\
        \quad \begin{tabular}[t]{@{}l@{}}
            $(1 - \text{normalized } Net\_delay^{s_i}_{c_j})$ \\
            ${}+ (\text{normalized } \phi^{r_n}_j)$ \\
            ${}+ (1 - \text{normalized } p^{c_j}_{s_i})$
        \end{tabular}
    \ENDFOR

    \STATE $SC: A^{r_n+1} \leftarrow Execute TTC(P^{r_n}, Q^{r_n}, Cap)$

    \STATE \textbf{PHASE C: System Update \& Next Round Preparation}
    \STATE Distribute new assignments $A^{r_n+1}$ to all nodes
    \STATE Distribute $\theta_{global}$ and neighbor weights to all $c_j \in C$

    \FORALL{client $c_j \in C$ in parallel}
        \STATE Update model with neighbors; update $\phi^{r_n}_j$ \& compute OR-based earnings
    \ENDFOR

    \STATE $r_n \leftarrow r_n + 1$
\ENDWHILE
\STATE \textbf{Return} $\theta_{global},\, A,\, \phi$
\end{algorithmic}
\end{algorithm}

\begin{algorithm}[H]
\caption{TTC Optimal Assignment Smart Contract}
\label{alg:TTC Optimal Assignment Smart Contract}
\begin{algorithmic}[1]
\REQUIRE client preferences $P$, Server preferences $Q$, Capacities $Cap$
\ENSURE Assignment mapping $A$
\STATE Initialize directed graph $G$ with nodes $C \cup S$
\STATE Initialize assignment map $A \leftarrow \emptyset$
\STATE Initialize preference queues: $queue_C, queue_S \leftarrow 0$
\FORALL{client $c_j \in C$}
    \STATE $top\_server \leftarrow P[c_j][queue_C[c_j]]$
    \STATE $G.AddEdge(c_j, top\_server)$
\ENDFOR
\FORALL{server $s_i \in S$}
    \STATE $top\_client \leftarrow Q[s_i][queue_S[s_i]]$
    \STATE $G.AddEdge(s_i, top\_client)$
\ENDFOR
\STATE $iteration \leftarrow 0$
\STATE \textbf{while} $G.HasNodes()$ \textbf{and} \\
\quad \begin{tabular}[t]{@{}l@{}}
    $iteration < MAX\_ITERATIONS$ \textbf{do}
\end{tabular}
    \STATE $cycle \leftarrow FindCycle(G)$
    \IF{$cycle = \emptyset$}
        \STATE \textbf{break}
    \ENDIF
    
    \FORALL{node $v \in cycle$}
        \IF{$v \in C$}
            \STATE $assigned\_server \leftarrow G.GetNextNode(v)$
            \STATE $A[v] \leftarrow assigned\_server$
            \STATE $current\_capacity \leftarrow$ \\
            \quad \begin{tabular}[t]{@{}l@{}}
                $|GetAssignedClients(assigned\_server)|$
            \end{tabular}
            \IF{$current\_capacity = Cap[assigned\_server]$}
                \STATE $G.RemoveNode(assigned\_server)$
            \ENDIF
            \STATE $G.RemoveNode(v)$
        \ENDIF
    \ENDFOR
    \STATE $UpdatePreferenceQueues(G, P, Q,$ \\
    \quad \begin{tabular}[t]{@{}l@{}}
        $queue_C, queue_S)$
    \end{tabular}
    \STATE $iteration \leftarrow iteration + 1$
\STATE \textbf{end while}
\STATE \textbf{return} $A$
\end{algorithmic}
\end{algorithm}

Third, a private Ethereum network, deployed via Kurtosis, addresses the high gas fees and latency associated with public Ethereum by allowing direct configuration of chain parameters, including the ability to set gas prices to zero, thereby removing cost barriers for frequent smart contract interactions \cite{Kurtosis}. In this work, we used the Polygon testnet, though any Ethereum-compatible testnet could be used interchangeably.

\subsection{Properties of SCOPE-FL}
The TTC mechanism has been known for its efficiency. In the original TTC, the preferences of students over schools and vice versa are fixed. However, in SCOPE-FL, these preferences can change during each FL round. In the subsequent theorem, we establish that SCOPE-FL maintains its efficiency even when preferences are variable. 

\textbf{Theorem 1.} SCOPE-FL produces an efficient assignment. \label{thm:1}

\medskip\noindent\textit{Proof.} In each FL round r, SCOPE-FL updates:

\begin{itemize}
\item The clients' preference profile P$^r$ over servers, and
\item The servers' priority profile Q$^r$ over clients (with capacities Cap),
\end{itemize}

then it runs the TTC mechanism on the resulting school-choice instance SC$^r$ = (C, S, Cap, P$^r$, Q$^r$). By the classic result of Abdulkadiroğlu--Sönmez, TTC returns a pareto efficient assignment for any fixed preferences/priorities with quotas \cite{abdulkadirouglu2003school}. Therefore, the assignment A$^{r}$ generated by SCOPE-FL in round r is pareto efficient in terms of (Cap, P$^r$, Q$^r$).

Dynamic evolution of preferences across rounds does not affect per-round efficiency: efficiency is evaluated relative to the preferences/priorities used in that round. Hence, SCOPE-FL maintains efficiency at every round by re-running TTC on the (new) fixed profiles.

\textbf{Theorem 2.} In SCOPE-FL, truthful preference revelation is a dominant strategy for all workers. No worker can
benefit by misrepresenting their preferences over servers.
(strategy proofness).

\medskip\noindent\textit{Proof.} According to~\cite{abdulkadirouglu2003school}, understanding why TTC is strategy-proof is intuitive. Consider a student matched at step $k$ under truthful reporting. At each preceding step, she pointed to her most preferred available school, meaning all schools she values more highly were already removed before reaching step $k$. Crucially, lying about her preferences cannot influence which cycles form in steps $1$ through $k-1$, so those preferred schools remain unattainable regardless. Consequently, misrepresentation offers no benefit and may only worsen her assignment.

To see that dynamic preference updates across rounds do not undermine strategy proofness, observe the following. In SCOPE-FL, preferences $P^r$ and priorities $Q^r$ are updated at the beginning of each round $r$ based on observed contribution scores $\phi^{r-1}_j$, negotiated prices, and network delays (all of which are computed prior to and independently of the TTC execution in that round). Consequently, within round $r$, the profiles $(P^r, Q^r)$ are fixed at the moment TTC is invoked, reducing the setting to a standard static school choice instance $\text{SC}^r = (C, S, \text{Cap}, P^r, Q^r)$. The classical strategy proofness result of Abdulkadiro\u{g}lu--S\"{o}nmez~\cite{abdulkadirouglu2003school} applies directly to this fixed instance: no worker can benefit by misreporting $P^r$ given $(Q^r, \text{Cap})$.

\section{Evaluation} \label{Evaluation}
To comprehensively evaluate SCOPE-FL's performance, we conducted simulation-based experiments assessing operational metrics including accuracy, reward distribution, latency, and convergence rate. Our evaluation framework compares it against baseline methods across all these parameters. The following sections detail our experimental methodology and present the simulation results.

\subsection{Experimental Setup}
Our simulation environment consisted of 10 servers, each with a capacity of 4 clients, and 50 participating clients. We set a uniform capacity of 4 clients per server as a practical experimental constraint to enable manageable computational overhead while ensuring sufficient diversity in matching outcomes. With 10 servers and 50 clients, this configuration allows for meaningful competition among clients for servers while maintaining tractable computation for the OR contribution evaluation method.

It is important to distinguish between two dimensions of scalability in SCOPE-FL: mechanism scalability and FL accuracy evaluation. The scalability of TTC, DA, and IAS is evaluated extensively in Sections ~\ref{subsec:Gas} and ~\ref{subsec:exetime}, where gas consumption and execution time are measured across configurations at 60×60×60 and 5000×5000×5000 respectively, demonstrating TTC's favorable scaling behavior. The FL accuracy experiments use a fixed configuration of 10 servers and 50 clients, intentionally designed to isolate matching quality rather than system throughput that is consistent with methodology in related works, such as HierMo, MAAIM, Spyker, and Block-RACS, which employ similarly sized client populations for mechanism comparison \cite{yang2023hierarchical,yellampalli2024client,zuo2024spyker, batool2023block}.

We evaluated the model's accuracy performance on three widely-used federated learning benchmark datasets: 
\begin{enumerate}
    \item CIFAR-10, a collection of 60,000 32×32 color images across 10 object classes \cite{Cifar-10}.
    \item Fashion-MNIST (F-MNIST), comprising 70,000 grayscale images of fashion items in 10 categories \cite{xiao2017fashion}.
    \item MNIST, containing 70,000 handwritten digit images (0-9) \cite{xiao2017fashion}.
\end{enumerate}

These datasets vary in complexity and are standard benchmarks for assessing federated learning systems. To evaluate SCOPE-FL under realistic federated learning conditions, we distributed data across clients using a heterogeneous (non-IID) partitioning strategy based on the Dirichlet distribution $\text{Dir}(\alpha)$, following widely adopted methodology in federated learning literature. The concentration parameter $\alpha$ controls the degree of heterogeneity: smaller $\alpha$ produces more skewed label distributions, while $\alpha \rightarrow \infty$ recovers the IID setting. In our primary experiments, we set $\alpha = 1.0$, which introduces mild label heterogeneity across clients, reflecting practical scenarios where client data distributions vary but are not drastically skewed.
{\sloppy

Dataset-specific CNN architectures were employed to balance performance and computational efficiency. For MNIST and Fashion-MNIST, we used a lightweight CNN with two convolutional layers (32, 64 filters), max-pooling, and two fully connected layers (128, 10 neurons) with 0.5 dropout, optimized via Adam. For CIFAR-10, we implemented a deeper architecture with four convolutional blocks (64, 128, 256, 512 filters) incorporating batch normalization, ReLU activations, max-pooling (first three blocks), and adaptive average pooling (final block). The classifier comprises three fully connected layers (512, 256, 10 neurons) with 0.3 dropout, trained using SGD with momentum (0.9) and adaptive learning rate scheduling.
\par}

The simulation framework was implemented in Python 3.x using PyTorch (deep learning framework), NumPy (numerical computations), and Python's built-in socket library for client-server communication. Model training leveraged PyTorch's neural network modules (torch.nn) and optimization algorithms (torch.optim), while data handling utilized torchvision for dataset loading and preprocessing. Inter-process communication between clients and servers was implemented using TCP/IP sockets with custom serialization protocols based on Python's pickle module.

Furthermore, the matching and assignment algorithms, namely TTC, DA, and IAS were implemented as the smart contract on the Ethereum blockchain using Solidity version 0.8.19. The development, deployment, and testing were conducted using Hardhat, a comprehensive Ethereum development environment. Interaction between the federated learning system and the blockchain was facilitated through Web3.py, a Python library that provides seamless integration with Ethereum nodes and enables programmatic interaction with smart contracts. Moreover, we tried to use gas-optimized operations in our smart contracts, such as the use of storage versus memory variables to minimize transaction costs.

Finally, we implemented SCOPE-FL on a 2016 MacBook Pro with a 2.2 GHz Quad-Core Intel Core i7, and 32 GB Memory and MacOS Monterey, version 12.7.6. Since all clients and servers were simulated on a single machine, the network delay term $Net\_delay^{s_i}_{c_j}$ does not correspond to physically measured round-trip times. Instead, we generated a synthetic latency profile at the start of each experiment, and an RTT value was drawn independently for every client-server pair for emulating the heterogeneous wide-area conditions of a geo-distributed HFL deployment. These values were used identically by all matching mechanisms (TTC, DA, IAS, MAAIM, and random selection).

\subsection{Baseline Methods}
To evaluate the effectiveness of SCOPE-FL, we conducted a comparison with two other algorithms, called DA and IAS; a modified version of a framework called MAAIM; and finally, a random selection mechanism. 

The DA algorithm, proposed by Gale and Shapley (1962), is a centralized matching mechanism that operates through iterative proposal and acceptance rounds. In each round, unmatched students propose to their most preferred school among those they have not yet applied to. Schools tentatively accept their most preferred applicants up to their capacity, rejecting others. Rejected students proceed to propose to their next preferred schools in subsequent rounds. The algorithm terminates when no student wishes to make further proposals. The resulting matching is stable. DA is strategy-proof only for the proposing side, and it doesn’t lead to a pareto efficient outcome. We applied the One-Round Reconstruction contribution evaluation method similar to SCOPE-FL and implement it on a smart contract like TTC \cite{roth2008deferred}.

The IAS mechanism operates through sequential application rounds where decisions are made immediately and irrevocably. In each round, unmatched students apply to their most preferred school that has not previously rejected them. Upon receiving applications, schools immediately accept their most preferred applicants up to available capacity and permanently reject others. Rejected students skip the rejecting school in subsequent rounds, removing it from consideration, and proceed to apply to their next preferred available school. The process continues until either all students are matched or no further applications are possible. Unlike DA, IAS makes binding decisions immediately rather than holding tentative acceptances, which can lead to unstable outcomes where blocking pairs exist. However, IAS offers computational simplicity and PE while the mechanism is not strategy-proof, as agents can benefit from misreporting preferences \cite{harless2014school}. 

In this method, we used influence function introduced in Block-RACS \cite{batool2023block}. The influence function serves as an alternative method to evaluate client contributions in federated learning by measuring the impact of individual clients on the global model's predictions. To calculate a client's influence, the global model is retrained by excluding that specific client, and the resulting change in model predictions is quantified. Formally, the influence of client $i$ is computed as the average absolute difference between predictions made by the complete global model (trained on all client data) and the retrained model (trained without client $i$), expressed as: $\text{Influence}_{i} = (1/n)\sum_{j=1}^{n} \widehat{W}_{n} - \widehat{W}_{n}^{-i}$, where n is the dataset size, $\widehat{W}_n$ represents the global model trained on all data, and $\widehat{W}_n^{-i}$ denotes the model trained excluding the i-th client \cite{batool2023block}. Since both IAS and influence function-based contribution assessment are computationally lightweight relative to other methods, we proposed a hybrid approach that combines these two mechanisms to investigate whether their integration can achieve competitive model performance while maintaining minimal computational overhead. Note that, like DA, we implemented IAS on a smart contract.

The MAAIM framework that employs a learning quality estimation approach based on loss reduction to guide server preferences in federated learning client selection. Specifically, they quantify each client's learning quality as the difference between the server's average global model loss in round $t$ and the client's local model loss in round $t+1$, expressed as: $q_i^t = loss_j(t) - loss_i^j(t + 1)$. This metric reflects the marginal improvement a client can provide to the global model, with higher values indicating greater learning utility. The authors
then utilize a many-to-one DA algorithm where servers rank clients based on descending learning quality values, while clients rank servers based on net monetary gain (server bid price minus energy consumption costs) \cite{yellampalli2024client}. It should be noted that our implementation of MAAIM deviates from the original in one respect: we exclude the energy consumption parameter from the client objective function, as it falls outside the scope of our experimental setup. Instead, we substitute latency as the communication cost proxy, consistent with the criteria used across all other baselines in this work. This adaptation ensures a fair and consistent comparison across all mechanisms. Therefore, whenever we use MAAIM, it refers to the modified version of this work.

Finally, random selection serves as a lower-bound baseline, 
where in each federated learning round, each tier-2 server randomly selects clients from the available pool up to its capacity $\text{Cap}_{s_i}$, without considering any preference, priority, contribution score, or incentive criterion. No matching mechanism or contribution evaluation method is employed. This baseline provides a reference point to quantify the minimum benefit that any structured 
client selection mechanism should surpass, and to contextualize the gains achieved by SCOPE-FL's pareto efficient matching approach. Table \ref{tab:comparison} summarizes how SCOPE-FL compares to existing approaches across the key mechanism design properties.

\begin{table*}[t]
\centering
\caption{Comparison of Mechanism Design Properties Across Client Selection Frameworks}
\label{tab:comparison}
\renewcommand{\arraystretch}{1.3}
\begin{tabular}{lccccc}
\hline
\textbf{Property} & \textbf{SCOPE-FL} & \textbf{DA} & \textbf{IAS} & \textbf{MAAIM} & \textbf{Random}\\
\hline
Pareto-Efficiency       & \checkmark         & \texttimes         & \checkmark         & \texttimes         & \texttimes \\
Stability             & Minimally unstable            & \checkmark         & \texttimes         & \checkmark         & \texttimes \\
strategy proofness      & \checkmark          & \checkmark      & \texttimes         & \checkmark      & \texttimes \\
Contribution Evaluation & Shapley-based (OR) & Shapley-based (OR) & Influence Function & Learning Quality   & \texttimes \\
Decentralized Execution & \checkmark         & \checkmark         & \checkmark         & \texttimes         & \texttimes \\
\hline
\end{tabular}
\end{table*}

\subsection{Evaluation Results}

This section evaluated SCOPE-FL (TTC) through comprehensive experiments across multiple performance dimensions. We organized our evaluation as follows: First, we assessed global model accuracy improvements on three standard FL datasets (MNIST, Fashion-MNIST, and CIFAR-10) compared to baseline methods. Second, we evaluated convergence efficiency of SCOPE-FL and TTC. Third, we analyzed average reward distribution in SCOPE-FL versus alternative baselines. Fourth, we examined the gas consumption overhead of implementing the TTC matching algorithm on EVM-based blockchains comparing with DA and IAS. Fifth, we measured the computational efficiency through algorithm running time analysis. Finally sixth, we evaluated average communication latency between matched client-server pairs in all methods.

\subsubsection{Accuracy}

We evaluated the global model accuracy achieved by SCOPE-FL (TTC) against four baseline client selection mechanisms that are explained in the Baseline Methods section across three benchmark federated learning datasets. Figure \ref{FIG:Accuracy results of SCOPE-FL compared to the baseline methods on CIFAR-10} illustrates the results on the CIFAR-10 dataset, being the most complex among the three benchmarks with 32×32 color images across 10 classes. As it is demonstrated in Figure \ref{FIG:Accuracy results of SCOPE-FL compared to the baseline methods on CIFAR-10}, after almost 40 FL rounds, SCOPE-FL with TTC achieves approximately 76\% global accuracy and significantly outperforms all baseline methods. DA reaches approximately 73\% accuracy, demonstrating reasonable but inferior convergence compared to SCOPE-FL. The gap is particularly notable in the early training rounds. IAS with influence function and MAAIM ranked third and fourth with 63\% and 48\%, respectively. random selection ranks last with approximately 37\% accuracy, confirming that unstructured client selection without any preference or contribution-aware mechanism fails to achieve competitive model quality on complex datasets such as CIFAR-10.

The accuracy convergence results over 30 training rounds on the Fashion-MNIST dataset are shown in Figure \ref{FIG:Accuracy results of SCOPE-FL compared to the baseline methods on F-MNIST}. Even though DA performs better than SCOPE-FL (TTC) in the early rounds, SCOPE-FL (TTC) shows better long-term convergence, ranking first out of all methods and reaching an accuracy of about 85\% after 30 rounds. With an accuracy of about 80\% and just 5\% less than SCOPE-FL (TTC), DA comes in second place. IAS comes in third place with about 78\% accuracy after 30 rounds, while MAAIM performs poorly, reaching only 72\% accuracy at the end of training. Random selection ranks last with approximately 66\% accuracy, further demonstrating that random client assignment provides insufficient model quality even on relatively simpler datasets.

\begin{figure}
	\centering
		\includegraphics[width=\columnwidth]{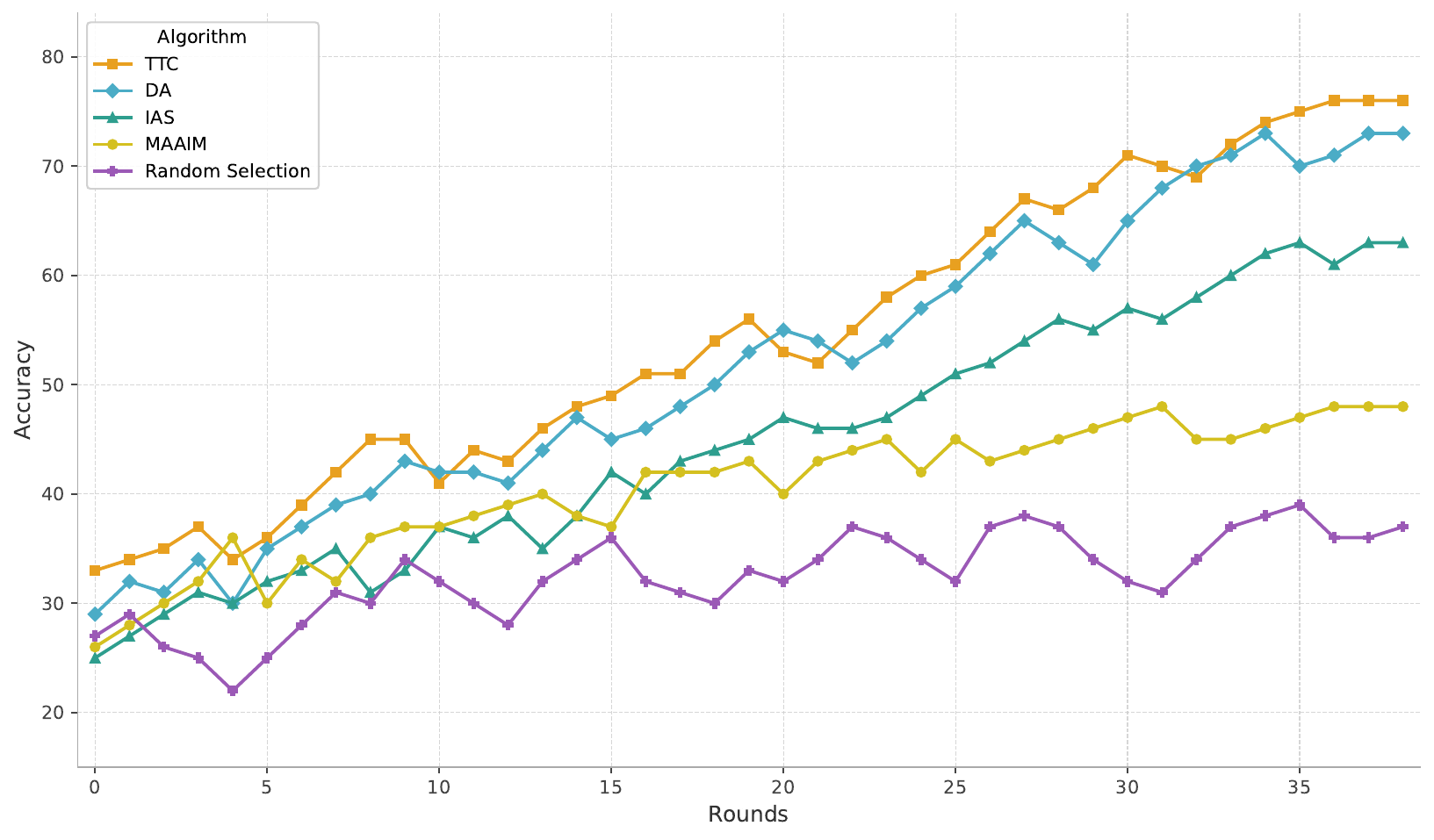}
	\caption{Accuracy Results of SCOPE-FL Compared to The Baseline Methods on CIFAR-10}
	\label{FIG:Accuracy results of SCOPE-FL compared to the baseline methods on CIFAR-10}
\end{figure}

\begin{figure}
	\centering
		\includegraphics[width=\columnwidth]{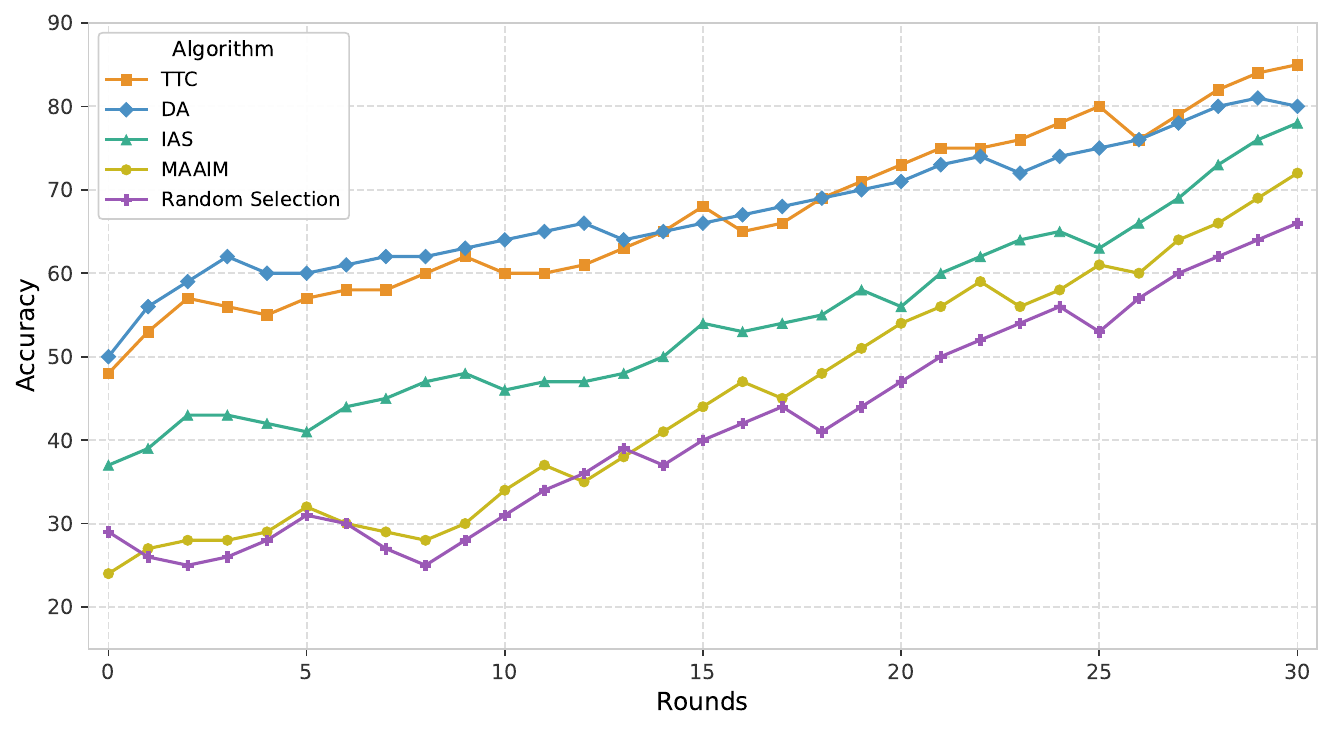}
	\caption{Accuracy Results of SCOPE-FL Compared to The Baseline Methods on F-MNIST}
	\label{FIG:Accuracy results of SCOPE-FL compared to the baseline methods on F-MNIST}
\end{figure}

Finally, Figure \ref{FIG:Accuracy results of SCOPE-FL compared to the baseline methods on MNIST} illustrates the accuracy results on the MNIST dataset over 18 FL rounds. SCOPE-FL (TTC) demonstrates superior performance, achieving approximately 95\% accuracy by round 18, while maintaining stable convergence throughout the training process. DA reaches approximately 92\% accuracy by the same round, showing slightly slower convergence in the initial rounds. IAS baseline ranks third with 90\% and more volatile behavior during the rounds, and MAAIM ranks fourth with almost 86\% accuracy and some fluctuations. Random selection ranks last with approximately 80\% accuracy, exhibiting the most unstable convergence behavior across all rounds, which is expected given the absence of any structured selection criterion.

\begin{figure}
	\centering
		\includegraphics[width=\columnwidth]{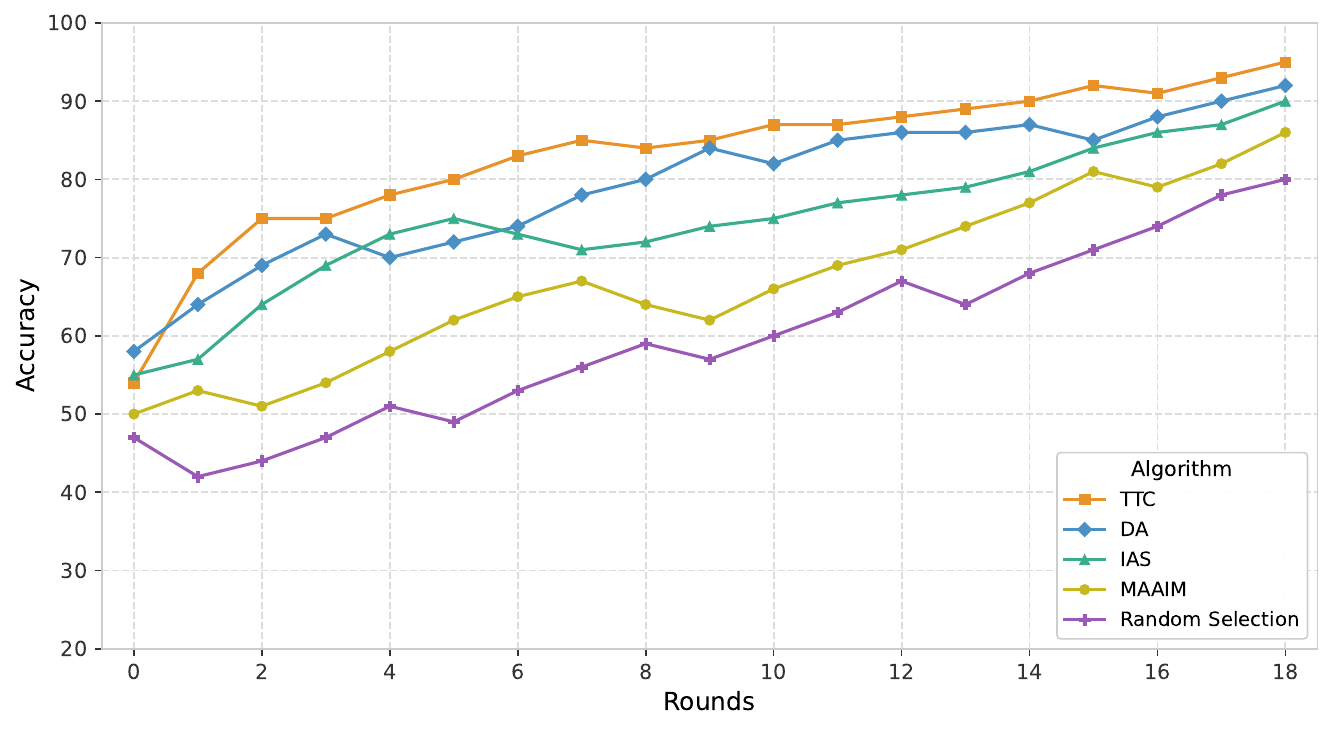}
	\caption{Accuracy Results of SCOPE-FL Compared to The Baseline Methods on MNIST}
	\label{FIG:Accuracy results of SCOPE-FL compared to the baseline methods on MNIST}
\end{figure}

The accuracy results across three federated learning datasets demonstrate that combining pareto efficient and minimally unstable mechanisms, such as TTC with the OR method for evaluating client contributions yields superior performance in the student-proposing school choice configuration for hierarchical and synchronous FL systems. Crucially, this approach also guarantees SP, which is essential in collaborative environments.

Notably, the performance gap between DA with OR and SCOPE-FL (TTC) with the same method is marginal across all three datasets. This narrow difference highlights DA's effectiveness in the mentioned setting. Nevertheless, achieving optimal accuracy requires mechanisms that balance both PE and minimal instability, such as TTC, paired with robust contribution evaluation methods like OR.

This balance is further evidenced by IAS, another pareto efficient mechanism combined with the influence function method, which consistently ranks third across all three datasets. These results emphasize that maximizing accuracy in federated learning requires the convergence of three key elements: PE, minimum instability, and accurate contribution evaluation. 

Finally, across all three datasets, random selection consistently ranks last, establishing a clear lower bound and confirming that even simple structured mechanisms, such as MAAIM, substantially outperform unguided client selection.

\subsubsection{Convergence Efficiency}

To further quantify SCOPE-FL's advantage, Figure~\ref{Fig:Cumulative Rounds Saved by TTC vs Baselines} illustrates the number of communication rounds saved by it relative to each baseline at six target accuracy levels per dataset. Across all three benchmarks, SCOPE-FL (TTC) consistently converges at least as fast as every competing mechanism, and its lead widens sharply as the target accuracy increases. On MNIST, SCOPE-FL saves up to 5 rounds over DA, 9 rounds over IAS, and 13 rounds over MAAIM, while random selection saves as many as 15 rounds and either fails to reach the two highest targets (85\% and 90\%) or requires a prohibitively large number of additional rounds to do so. On Fashion-MNIST, the advantage becomes more pronounced at higher targets: relative to DA, IAS, and MAAIM the savings grow to 8, 13, and 17 rounds respectively, while MAAIM and random selection either fail to reach the most demanding accuracy thresholds or converge to them only after a long training horizon. On the more challenging CIFAR-10 dataset, SCOPE-FL saves up to 2 rounds over DA and 15 rounds over IAS, while MAAIM either fails to attain the three highest targets or reaches them prohibitively slowly, and random selection reaches only the lowest target accuracy of 40\%, requiring 15 additional rounds compared to it. Notably, the gap against MAAIM grows so steeply with target difficulty that on CIFAR-10 and Fashion-MNIST it either does not converge to the harder targets at all or does so only after an excessive number of rounds, reflecting MAAIM's slow convergence due to its learning-quality estimation method. The marginal advantage over DA remains comparatively small but consistent, reinforcing the earlier observation that DA is a competitive baseline when paired with OR-based contribution evaluation. These results collectively demonstrate that SCOPE-FL's pareto efficient matching not only achieves higher final accuracy but also reaches target performance thresholds significantly faster, reducing the overall energy consumption and communication overhead of the federated learning process.

\begin{figure*}[t]  % t=top of page, b=bottom of page
    \centering
    \includegraphics[width=1\textwidth]{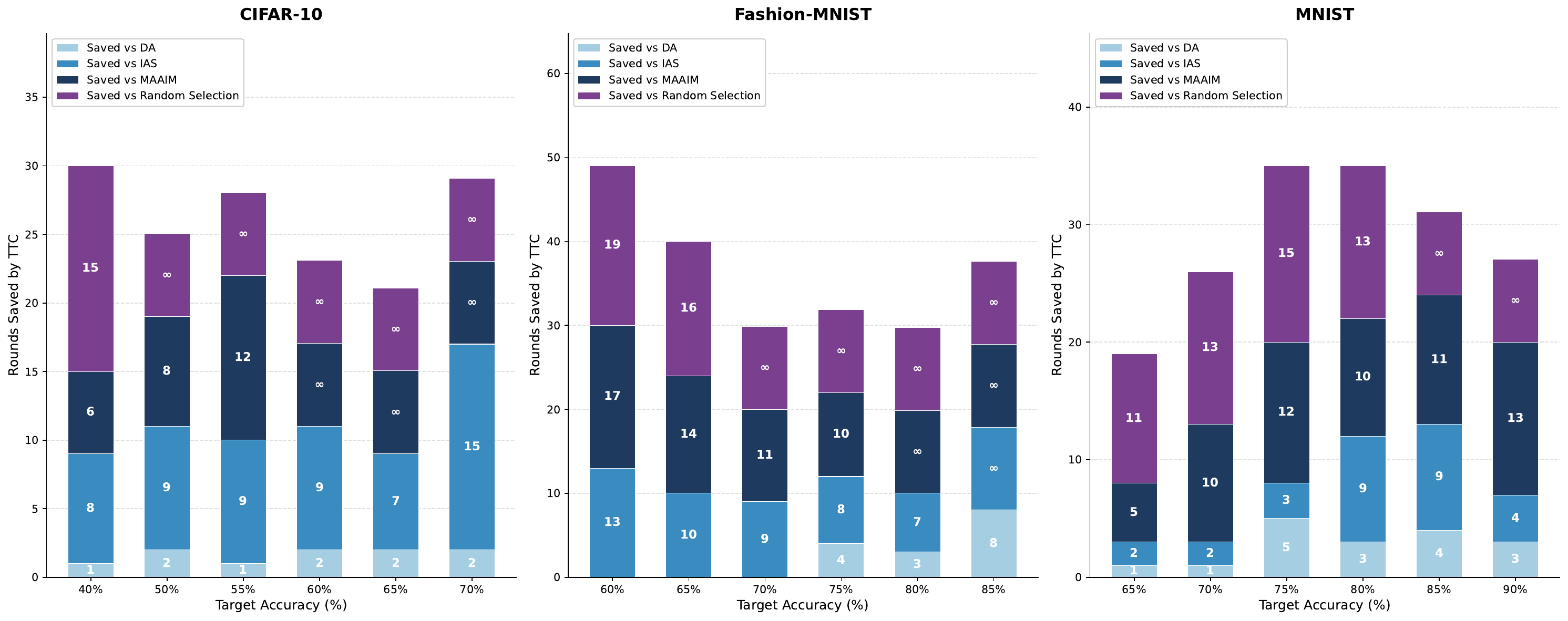}
    \caption{Cumulative Rounds Saved by TTC vs Baselines}
    \label{Fig:Cumulative Rounds Saved by TTC vs Baselines}
\end{figure*}

\subsubsection{Reward}
Figure \ref{FIG:Average paid reward by servers across rounds for different mechanisms} presents the average obtained reward by clients across rounds for different methods. From the clients' perspective, TTC delivers the highest average reward received at approximately 69.40, with a relatively wide distribution ranging from 60 to 79. This result demonstrates TTC yields the most favorable client-side outcomes through substantial compensation.

DA ranks second in terms of rewards received at 64.05, showing favorable outcomes for clients with a median of 65 and a distribution ranging from 52 to 73. This aligns with theoretical expectations, as DA is optimal for clients (students) among stable matchings in the client-proposing variant. IAS with influence function follows at 59.35, displaying a consistent distribution with moderate client rewards.  MAAIM with learning quality estimation demonstrates the least favorable outcome for clients with the lowest average reward received of 49.60.

Random selection, as an unstructured baseline, exhibits the highest variance across all methods, with rewards ranging from 29 to 79 and a mean of approximately 46.80. Despite occasionally producing high rewards, its wide and unpredictable distribution reflects the absence of any strategic matching logic, making it unreliable from the clients' perspective.

The results highlight a fundamental trade-off in federated learning matching systems: while TTC with OR method delivers the highest client reward, MAAIM with learning quality estimation minimizes rewards received by clients, making it the least economically advantageous structured mechanism. Random selection, though occasionally competitive, fails to provide consistent compensation, confirming that unstructured assignment is unsuitable for client-welfare-oriented deployment. DA with OR method and IAS with influence function represent intermediate positions, with DA providing near-optimal client rewards while IAS offers moderate compensation. This demonstrates TTC's clear advantage in providing superior economic outcomes for clients, ensuring they receive maximum compensation for their contributions to the federated learning system.

\begin{figure}
	\centering
		\includegraphics[width=\columnwidth]{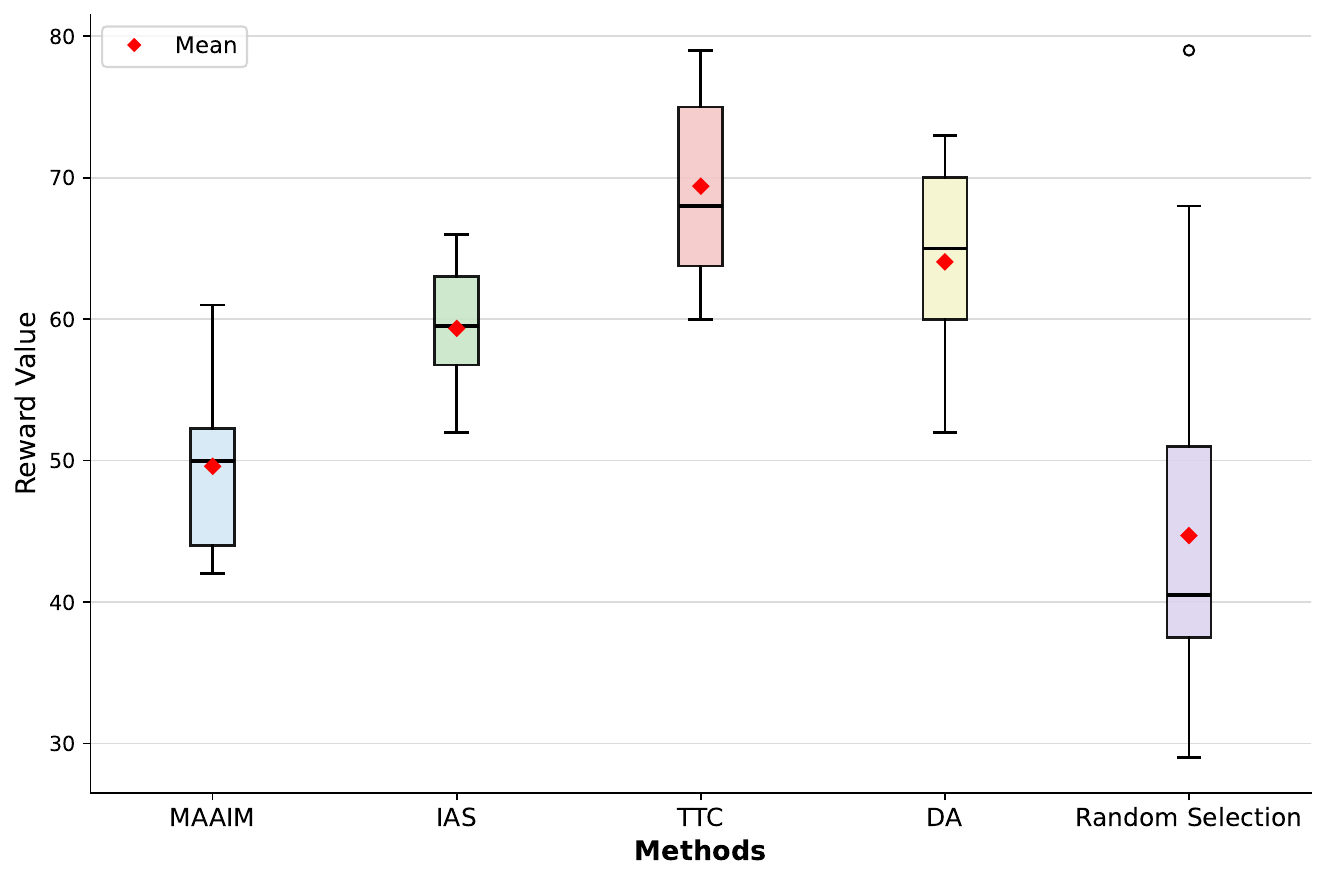}
	\caption{Average Obtained Reward by Clients Across Rounds for Different Mechanisms}
	\label{FIG:Average paid reward by servers across rounds for different mechanisms}
\end{figure}

\subsubsection{Latency}
Figure \ref{FIG:Average latency across rounds for different mechanisms} presents the average latency across rounds for different mechanisms. As mentioned earlier, clients assign a weight of 0.5 to latency, while servers assign 0.3 to this parameter. The results show that DA achieves the lowest average latency at approximately 588 ms, ranking first. The SCOPE-FL with TTC following closely at 609 ms (a marginal difference of only 21 ms). The boxplot further demonstrates that both DA and SCOPE-FL (TTC) maintain highly consistent performance, with DA exhibiting the tightest distribution and minimal variability. In contrast, IAS and MAAIM show substantially higher latencies with greater variability; MAAIM in particular displays the widest latency range, while IAS presents several outliers indicating inconsistent performance across rounds. Thus, while DA leads in latency optimization, SCOPE-FL (TTC) also demonstrates competitive and stable performance in this metric. Finally, random selection exhibits the highest average latency among all methods, with substantially greater variability across rounds, reflecting the absence of any latency-aware assignment criterion. Since clients and servers are paired arbitrarily, there is no mechanism to favor low-RTT pairs, resulting in consistently poor communication efficiency.

Note that DA's small latency advantage is expected, and its lower reward has the same cause. As the client-optimal stable mechanism, DA fully captures the low-RTT pairings that clients prefer, but it cannot do the same for reward. Latency is an aligned attribute, since both clients and servers seek to minimize $Net\_delay$, so a stable matching secures low-RTT pairs without conflict. Reward, however, is contested, because clients want a high price while servers want a low one. Stability therefore prevents DA from raising rewards in clients' favor without creating blocking pairs. As a result, DA optimizes the uncontested dimension (latency) fully but settles for a constrained outcome on the contested one (reward). TTC, which is not bound by stability, trades the marginal 21~ms (about 3.5\%) latency gap for the higher reward shown in Fig.~\ref{FIG:Average paid reward by servers across rounds for different mechanisms}.

Beyond per-round latency, SCOPE-FL's faster convergence directly reduces total communication overhead over the training lifetime since the rounds saved by SCOPE-FL (TTC) over each baseline translate directly into fewer rounds of model upload and download across all matched client-server pairs, representing a proportional reduction in aggregate bandwidth consumption.

\begin{figure}
	\centering
		\includegraphics[width=\columnwidth]{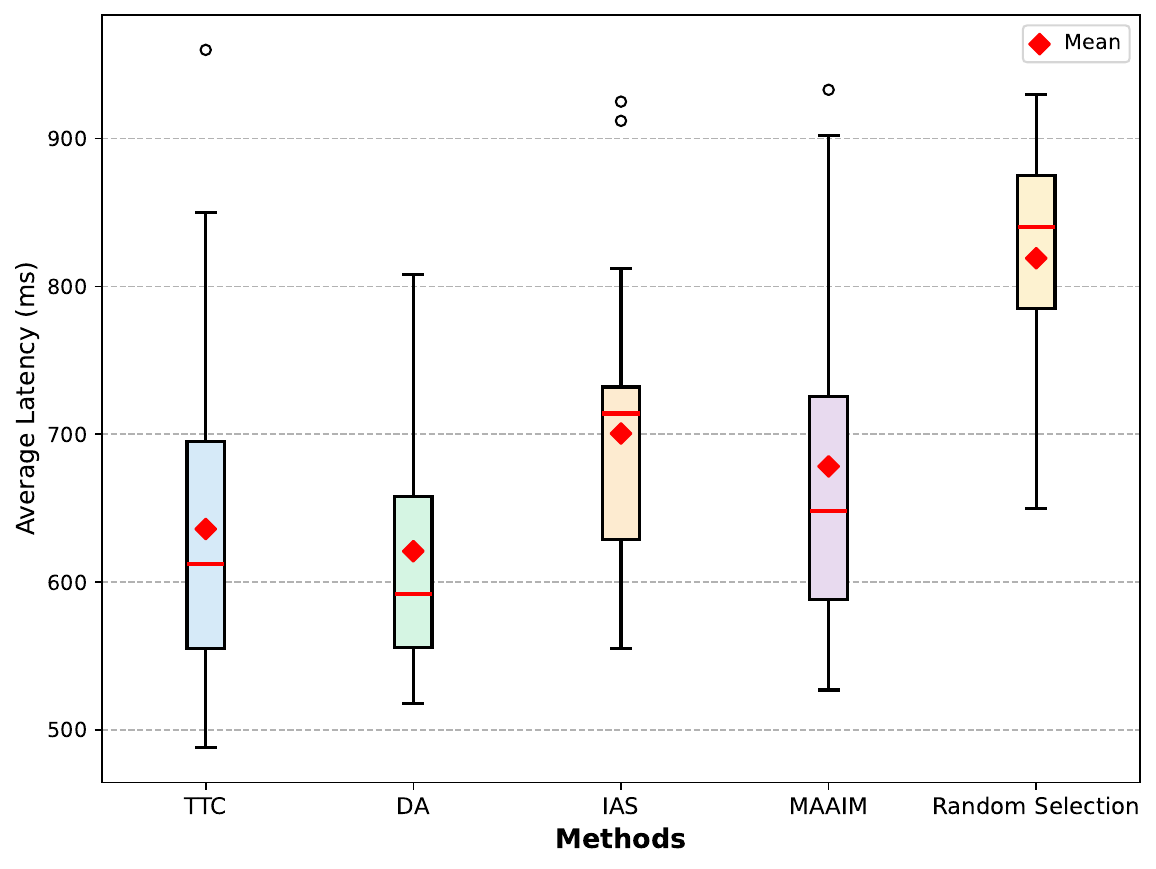}
	\caption{Average Latency Across Rounds for Different Mechanisms}
	\label{FIG:Average latency across rounds for different mechanisms}
\end{figure}
\subsubsection{Gas Consumption} \label{subsec:Gas}
Figure \ref{FIG:Gas Consumption Comparison between three mechanisms} shows the gas consumption comparison across three matching mechanisms, including DA, TTC, and IAS as the system scales from 5×5×5 to 60×60×60 configurations (Number of clients, number of servers, and number of preferences). The results reveal clear differences in efficiency among the mechanisms. At the smallest scale (5×5×5), all three mechanisms have similar and low gas consumption, below 10,000,000 units. However, as the system grows larger, the differences become more apparent.

TTC consistently shows the lowest gas consumption across all test cases. At the largest scale (60×60×60), TTC uses approximately 98,000,000 gas units. IAS performs similarly with about 104,000,000 gas units, showing only a small difference from TTC. DA, however, consumes significantly more gas, especially at larger scales. At the 30×30×30 configuration, DA uses about 76,000,000 gas units while TTC uses only 20,000,000 units. This difference grows much larger at the 60×60×60 scale, where DA reaches approximately 260,000,000 gas units (More than 2.5 times higher than TTC and IAS).

The steep increase in DA’s gas consumption shows it does not scale well for larger systems. In contrast, TTC and IAS show more gradual increases. These results demonstrate that TTC is not only effective in terms of matching quality and monetization, but also the most cost-efficient option for blockchain-based federated learning systems, particularly in large-scale deployments.

The gas efficiency advantage of TTC over DA stems from fundamental differences in storage write complexity on the EVM. On EVM-compatible blockchains, the SSTORE opcode (which writes a value to contract storage) costs 20,000 gas for writing a new value and 5,000 gas for updating an existing one, making it the single most expensive operation in smart contract execution \cite{ETHEREUM}. 

DA's tentative acceptance mechanism requires iterative proposal-rejection cycles where students may be accepted, displaced, and reassigned multiple times, resulting in O(n²) cumulative SSTORE operations in the worst case. Each time a school displaces a tentatively accepted student in favor of a higher-priority proposer, two storage writes occur: one to record the new acceptance and one to invalidate the previous one. IAS, while avoiding DA's tentative acceptance cycles, still incurs multiple SSTORE operations per worker in the worst case. Since IAS makes immediate and irrevocable decisions, each worker may be permanently rejected by every server except their final match, requiring a storage write for each rejection and preference pointer advancement. In the worst case, where every worker is rejected by all but their last preferred server, this results in $O(n^2)$ cumulative SSTORE operations, similar in complexity to DA but arising from a fundamentally different mechanism: rejection cascades rather than displacement cycles.

In contrast, TTC's cycle-clearing mechanism produces permanent, write-once assignments, meaning once a client-server pair is matched through a cycle, that assignment is final and never mutated, incurring exactly one 20,000-gas SSTORE write per matched pair. This yields O(n) storage writes in total. Since SSTORE operations dominate gas consumption on EVM-compatible blockchains, this difference in storage write complexity directly explains the empirical gas consumption gap observed in Figure \ref{FIG:Gas Consumption Comparison between three mechanisms}, where DA's consumption grows superlinearly compared to TTC's near-linear scaling.

Finally, random selection is excluded from this comparison as it involves no matching mechanism and incurs negligible on-chain overhead, including no graph construction, cycle detection, or iterative storage writes, making its gas cost effectively trivial regardless of scale.

\begin{figure}
	\centering
		\includegraphics[width=\columnwidth, height= 5cm]{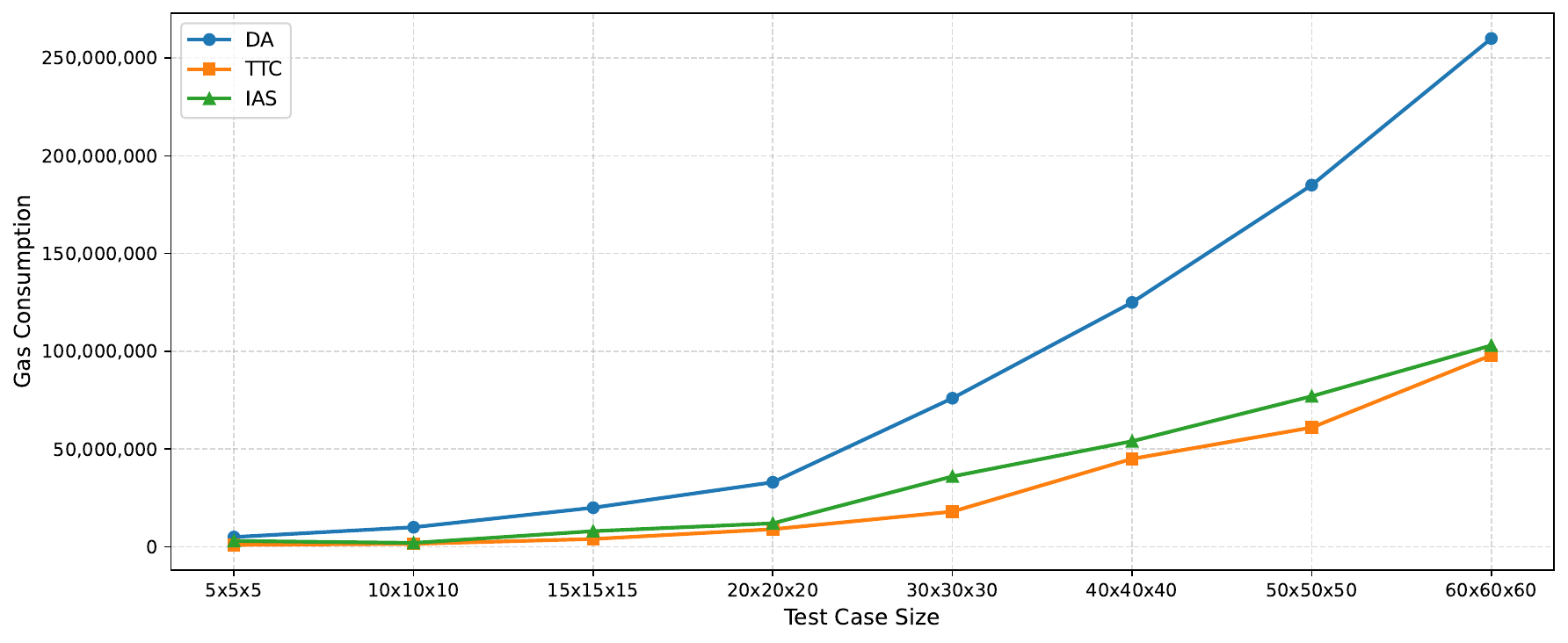}
	\caption{Gas Consumption Comparison Between Three Mechanisms}
	\label{FIG:Gas Consumption Comparison between three mechanisms}
\end{figure}

\subsubsection{Execution Time}\label{subsec:exetime}
Figure \ref{FIG:Execution Time of three mechanisms} presents the execution time comparison of DA, TTC, and IAS across test case sizes from 5$\times$5$\times$5 to 5000$\times$5000$\times$5000 (Number of clients, number of servers, and number of preferences).

At small scales (up to 500$\times$500$\times$500), all three algorithms show negligible execution times below 100 milliseconds. However, as the system scales, performance differences emerge. At the largest scale (5000$\times$5000$\times$5000), IAS demonstrates the best performance at approximately 5300 milliseconds, followed by TTC at 6300 milliseconds, and DA at 6500 milliseconds.

These differences stem from algorithmic complexity. DA's iterative proposal-rejection mechanism requires multiple processing rounds. TTC's cycle-finding approach is more efficient but requires complete graph traversal. IAS achieves the best execution time through its incremental adjustment strategy, which avoids redundant computations. The results show that while all mechanisms are practical for moderate-scale deployments, IAS offers the best scalability for large systems, followed by TTC, with DA being the most computationally intensive. Note that random selection is similarly omitted, as its O(1) per-server selection logic incurs near-zero computational cost regardless of scale, offering no meaningful basis for algorithmic comparison.
\begin{figure}
	\centering
		\includegraphics[width=\columnwidth, height = 5.5cm]{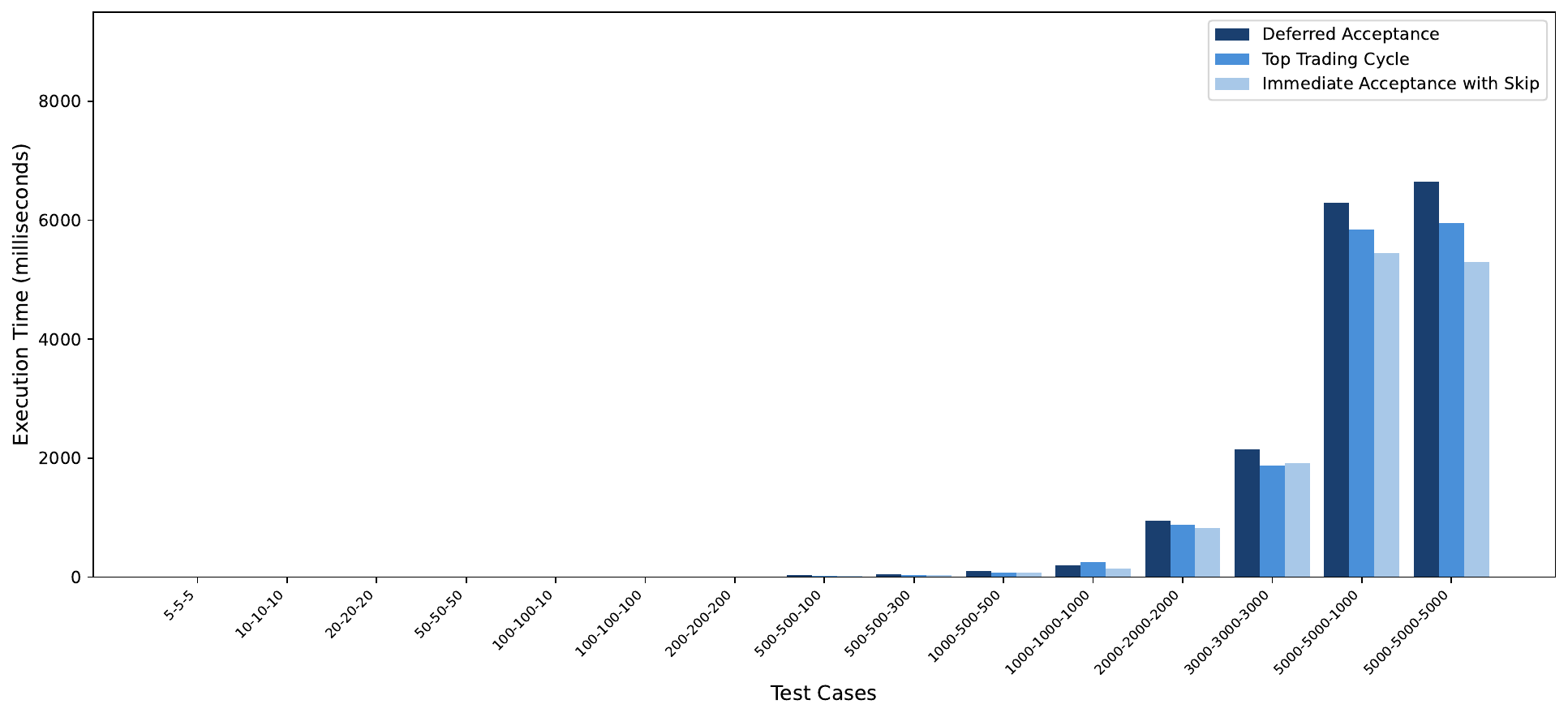}
	\caption{Execution Time of Three Mechanisms}
	\label{FIG:Execution Time of three mechanisms}
\end{figure}

\section{Conclusion} \label{Conc}
Client selection in hierarchical federated learning has long suffered from a fundamental strategic inefficiency: existing mechanisms either sacrifice PE for stability, or achieve efficiency without eliminating incentives for preference manipulation. Either failure degrades system welfare in practice.

SCOPE-FL addresses this directly by formulating client selection as a two-sided school choice problem solved through the Top Trading Cycle algorithm, simultaneously achieving PE and SP, a combination no prior HFL framework has provided. Contribution evaluation via a customized One-Round Reconstruction Shapley approximation for HFL systems ensures reward distribution accurately reflects each client's true marginal impact, while blockchain smart contract execution makes these guarantees tamper-proof in adversarial conditions.

Evaluation on MNIST, Fashion-MNIST, and CIFAR-10 demonstrates consistent outperformance over DA, IAS, and other baselines across accuracy, convergence rate, reward efficiency, and blockchain overhead with communication latency and mechanism execution time comparable to DA and IAS, respectively.

Future work includes exploration of flat multi-server architectures to eliminate hierarchical dependencies and improve scalability by allowing a larger and more diverse client population to participate without the overhead of tiered aggregation, and replacing FedAvg with FedProx as the local aggregation algorithm to better handle more intense non-IID data distributions across clients, where the proximal regularization term can stabilize local training under high statistical heterogeneity.

\bibliographystyle{IEEEtran}
\bibliography{main}

% \newpage
% \vfill

\clearpage

% Reset and prefix figure/table numbers for appendix
% \setcounter{figure}{0}
% \renewcommand{\thefigure}{A.\arabic{figure}}
% \setcounter{table}{0}
% \renewcommand{\thetable}{A.\arabic{table}}

% % Centered Appendix header
% \begin{center}
%     {\large\textbf{APPENDIX}}
% \end{center}

% \appendix

% \section{TTC Step-by-Step Example}
% \label{appendix:ttc_example}

{\appendix[TTC Step-by-Step Example]
\label{app:ttc_example}

To demonstrate the algorithm's steps, we will outline its procedure using an example. Consider five individuals, Alice, Bob, John, Lisa, and Suzanne, and five objects, A, B, C, D and E. Table \ref{tab:ttc_prefs} gives the preferences of the individuals over those objects and vice versa.

% Non-floating table — cannot drift away from this position
\begin{center}
\refstepcounter{table}
\textbf{Table~\thetable} \\
\small Objects' and agents' preferences and priorities~\cite{haeringer2018market}
\label{tab:ttc_prefs}
\vspace{0.3cm}

\begin{tabular}{ccccc}
$P_{Alice}$ & $P_{Bob}$ & $P_{John}$ & $P_{Lisa}$ & $P_{Suzanne}$ \\
\hline
B & C & A & C & A \\
E & A & E & A & C \\
D & D & D & E & B \\
C & E & B & B & D \\
A & B & C & D & E \\
\end{tabular}

\vspace{0.4cm}

\begin{tabular}{ccccc}
$P_A$ & $P_B$ & $P_C$ & $P_D$ & $P_E$ \\
\hline
Alice   & Bob     & John    & Lisa    & Bob     \\
John    & Lisa    & Suzanne & Suzanne & Suzanne \\
Bob     & Suzanne & Bob     & Alice   & Alice   \\
Suzanne & Alice   & Lisa    & John    & John    \\
Lisa    & John    & Alice   & Bob     & Lisa    \\
\end{tabular}
\end{center}

\vspace{0.3cm}

% Step 1 text and Figure A.1 side by side in a minipage
% This forces them to stay together in the same column block
\noindent
\begin{minipage}[t]{\columnwidth}
\textbf{Step 1}: For each individual, we create an arrow pointing to their most preferred object (for instance, Alice points to B, Bob to C, and John to A). Likewise, for every object we draw an arrow toward the individual with the highest priority for that object (e.g., object D points to Lisa, who holds the highest priority). From this, we can identify a cycle consisting of the solid arrows: (Alice,\,B;\;Bob,\,C;\;John,\,A). Arrows not included in any cycle are shown as dashed lines. All individuals and objects that are part of the cycle are immediately matched and removed from the system. Hence, the assignments at this stage are: $\mu(\text{Alice}) = B$, $\mu(\text{Bob}) = C$, $\mu(\text{John}) = A$. The resulting graph is shown in Figure~\ref{FIG:First step of TTC}.

\vspace{0.3cm}
\refstepcounter{figure}
\begin{center}
    \includegraphics[width=\columnwidth]{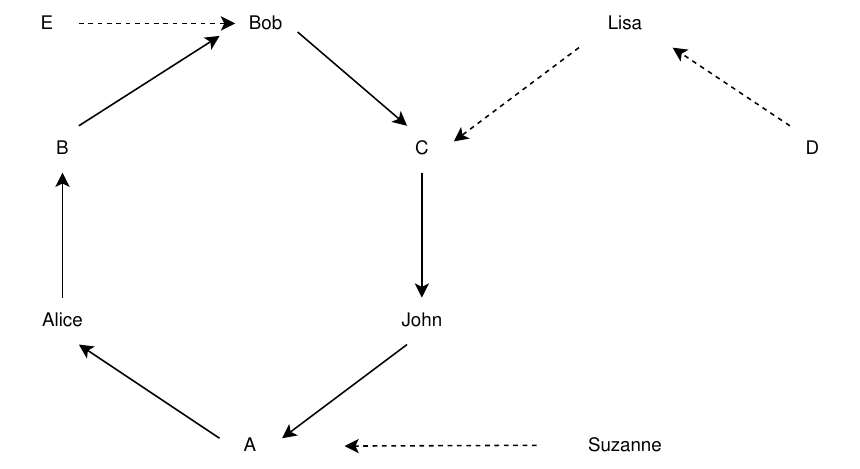}\\
    \vspace{0.2cm}
    \small\textbf{Figure~\thefigure}: First step of TTC
    \label{FIG:First step of TTC}
\end{center}
\end{minipage}

\vspace{0.4cm}

% Step 2 text and Figure A.2 side by side in a minipage
\noindent
\begin{minipage}[t]{\columnwidth}
\textbf{Step 2}: At this point, two individuals (Lisa and Suzanne) and two objects (D and E) remain. We now draw an arrow from each individual to her most preferred object among the remaining ones. Similarly, each object points to the individual with the highest priority between Lisa and Suzanne. Thus, object D points to Lisa, and object E points to Suzanne. There is a new cycle, so Lisa and Suzanne are assigned to the object they are pointing to, and since there is no individual who is left in the problem, the algorithm stops. The final assignment is: $\mu(\text{Alice}) = B$, $\mu(\text{Bob}) = C$, $\mu(\text{John}) = A$, $\mu(\text{Lisa}) = E$, $\mu(\text{Suzanne}) = D$. The corresponding graph is given in Figure~\ref{FIG:Second step of TTC}.

\vspace{0.3cm}
\refstepcounter{figure}
\begin{center}
    \includegraphics[width=0.7\columnwidth]{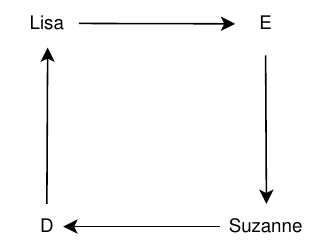}\\
    \vspace{0.2cm}
    \small\textbf{Figure~\thefigure}: Second step of TTC
    \label{FIG:Second step of TTC}
\end{center}
\end{minipage}}

\end{document}